\documentclass[10pt,twocolumn,letterpaper]{article}

\usepackage{cvpr}              %

\usepackage{graphicx}
\usepackage{amsmath}
\usepackage{amssymb}
\usepackage{booktabs}
\usepackage{arydshln}

\usepackage[pagebackref,breaklinks,colorlinks]{hyperref}

\usepackage{enumitem}

\usepackage[capitalize]{cleveref}
\crefname{section}{Sec.}{Secs.}
\Crefname{section}{Section}{Sections}
\Crefname{table}{Table}{Tables}
\crefname{table}{Tab.}{Tabs.}

\def\cvprPaperID{10574} %
\def\confName{CVPR}
\def\confYear{2022}

\newcommand{\squeezeup}{\vspace{-2mm}}

\usepackage[numbers]{natbib}

\usepackage{microtype}
\usepackage{graphicx}
\usepackage[space]{grffile}
\usepackage{booktabs} %

\usepackage{amsmath}
\usepackage{amssymb}
\usepackage{amsthm}
\usepackage{amsfonts}
\usepackage{mathtools}
\usepackage{bbm}
\usepackage{nicefrac}
\usepackage{lipsum}
\usepackage[labelfont=bf]{caption}
\usepackage{subcaption}
\usepackage{wrapfig}
\usepackage{xcolor}

\usepackage{multibib}

\usepackage[export]{adjustbox}

\graphicspath{ {images/} }

\def\argminop{\mathop{\rm arg\,min}\limits}

\newcommand{\sign}{\mathop{\rm sign}}
\def\R{\mathbb{R}}

\newcommand{\norm}[1]{\left\|#1\right\|}

\DeclareMathOperator{\eps}{\mathbf{\varepsilon}}
\DeclareMathOperator{\E}{\mathbb{E}}
\let\l\relax
\DeclareMathOperator{\l}{\mathbf{\ell}}

\newcommand{\comment}[1]{}

\renewcommand{\paragraph}{\textbf}  %
\setitemize{topsep=3.5pt,parsep=0pt,partopsep=0pt,leftmargin=25pt}

\usepackage{enumitem}

\makeatletter
\def\blfootnote{\xdef\@thefnmark{}\@footnotetext}
\makeatother

\begin{document}

\title{ARIA: Adversarially Robust Image Attribution for Content Provenance}

\author{
    Maksym Andriushchenko$^*$\\ 
    EPFL\\
    {\tt\small maksym.andriushchenko@epfl.ch}
    \and
    Xiaoyang Rebecca Li\\ 
    Adobe Research\\
    {\tt\small xiaoli@adobe.com}
    \and
    Geoffrey Oxholm\\ 
    Adobe Research\\
    {\tt\small oxholm@adobe.com}
    \and
    Thomas Gittings\\ 
    University of Surrey\\
    {\tt\small t.gittings@surrey.ac.uk}
    \and
    Tu Bui\\ 
    University of Surrey\\
    {\tt\small t.v.bui@surrey.ac.uk}
    \and
    Nicolas Flammarion\\ 
    EPFL\\
    {\tt\small nicolas.flammarion@epfl.ch}
    \and
    John Collomosse\\ 
    Adobe Research\\
    {\tt\small collomos@adobe.com}
}
\maketitle

\begin{abstract}
Image attribution -- matching an image back to a trusted source -- is an emerging tool in the fight against online misinformation. Deep visual fingerprinting models have recently been explored for this purpose. However, they are not robust to tiny input perturbations known as adversarial examples. First we illustrate how to generate valid adversarial images that can easily cause incorrect image attribution. Then we describe an approach to prevent imperceptible adversarial attacks on deep visual fingerprinting models, via robust contrastive learning. The proposed training procedure leverages training on $\ell_\infty$-bounded adversarial examples, it is conceptually simple and incurs only a small computational overhead. The resulting models are substantially more robust, are accurate even on unperturbed images, and perform well even over a database with millions of images. In particular, we achieve 91.6\% standard and 85.1\% adversarial recall under $\l_\infty$-bounded perturbations on manipulated images compared to 80.1\% and 0.0\% from prior work. We also show that robustness generalizes to other types of imperceptible perturbations unseen during training. Finally, we show how to train an adversarially robust image comparator model for detecting editorial changes in matched images.
\end{abstract}

\blfootnote{\vspace{1em} \textsuperscript{*}Work done as internship at Adobe Research.}

\begin{figure*}[t!]
    \centering
    \includegraphics[width=1.0\linewidth]{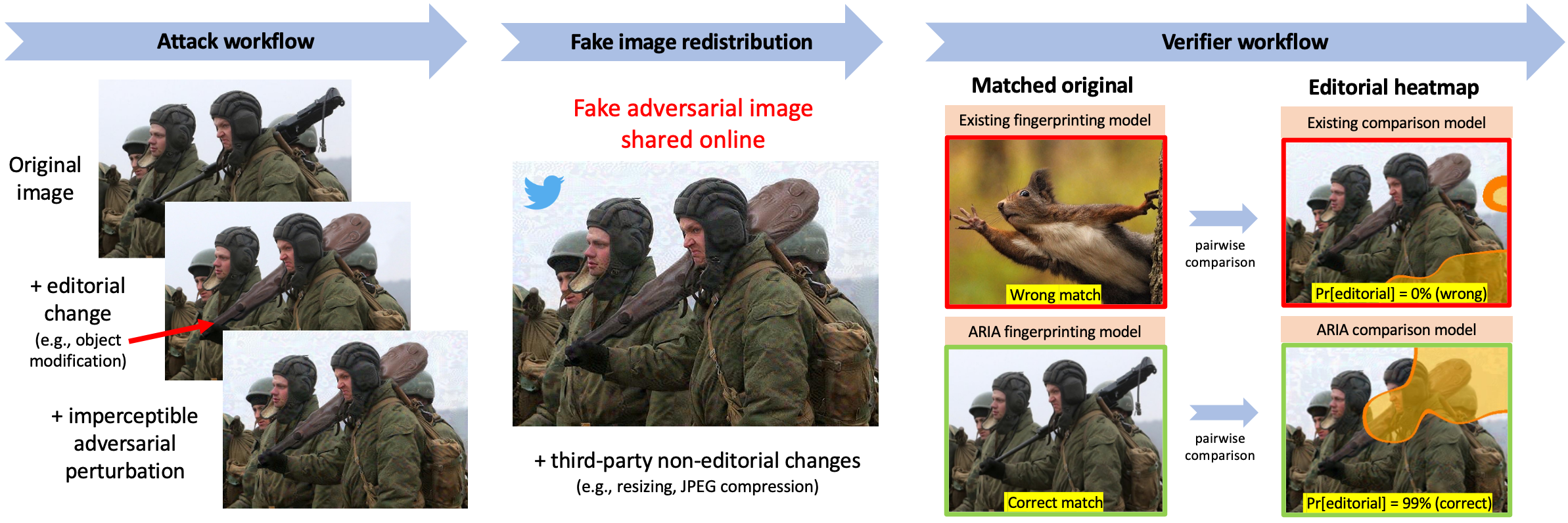}
     \caption{\textbf{Workflow of image attribution}. ARIA applied to the framework of \citet{black2021deep} enables both robust \textit{matching} and \textit{comparison} of manipulated images, despite the presence of imperceptible adversarial attacks. We consider an attack workflow in which an attacker adds both editorial changes (e.g., an object modification or a face swap) and an adversarial perturbation prior to the distribution of the adversarial image online which may involve additional non-editorial changes (e.g., resizing, JPEG compression). Existing image attribution models can be fooled by such attacks: the image fingerprinting model produces an irrelevant match and the comparison model predicts either no change or false editorial changes. At the same time, our ARIA training leads to highly accurate models that perform \textit{adversarially robust} matching and comparison.
     }
    \label{fig:teaser}
\end{figure*}

\section{Introduction}
Fake news and misinformation are major societal threats being addressed by new computer vision methods to  determine content authenticity. Such methods fall into two camps: detection and attribution.  Detection methods automatically identify manipulated or synthetic images through visual artifacts or statistics \cite{Zhang2020Attacks,zhang2019,wang2021deepfake}. Attribution methods match an image to a trusted database of originals \cite{nguyen2021oscar,black2021deep,blackcvmp}.  Once matched, any differences may be visualized, and any associated provenance data displayed. Rather than making automated judgments, the goal of image attribution is to enable users to make more informed trust decisions \cite{cai_article_witness_org}.

This paper considers specifically the {\em image attribution} problem where the goal is to differentiate between `non-editorial' transformation of content (\eg due to resolution, format or quality change) and editorial change where content is digitally altered to change its meaning.  \citet{nguyen2021oscar} use contrastive training to learn a visual hashing %
function that is invariant to non-editorial, but sensitive to editorial changes.  In such `tamper-sensitive' matching, a manipulated image would not be falsely corroborated by provenance data associated with the original.  By contrast, \citet{black2021deep} learn a `tamper-invariant' image fingerprint which is insensitive to {\em both} non-editorial and editorial change, and visually highlight manipulated changes using a separate model.

This paper reports on the novel problem of {\em adversarial attack and defense for these image attribution approaches}.   They rely on deep neural networks to learn visual fingerprints for near-duplicate image matching.  However, deep networks are known to be vulnerable to {\em adversarial attacks} \citep{szegedy2013intriguing} that use subtle image perturbations to cause dramatic changes in the output of the models.  Visual search models based on deep networks are not exceptions and recently have been shown to be also vulnerable to adversarial attacks \citep{dolhansky2021adversarial}.   

We make the following technical contributions:
\begin{enumerate}[label=\textbf{\arabic*.}, itemsep=1mm, leftmargin=0mm, labelwidth=2cm, itemindent=5mm, topsep=2mm]
    \item \textbf{Adversarial attack of image attribution models.}  
    We present a white-box gradient-based method for crafting adversarial examples to attack both tamper-sensitive and tamper-invariant image attribution models.  We show it is possible to closely match the perceptual fingerprints of unrelated images by using small $\l_\infty$-adversarial perturbations.  For tamper-sensitive models, we show that the original image may be incorrectly matched given a manipulated query.  For tamper-invariant models, we additionally show that the image comparison post-process used to visualize areas of image manipulation may also be fooled to show either no manipulation or false areas of manipulation. 
    Thus, we show that trust in both state-of-the-art image attribution approaches may be defeated by adversarial examples.
    
    \item \textbf{Robust contrastive learning for image attribution.} 
    We describe a novel robust contrastive learning algorithm to train image fingerprinting models for attribution, to ensure robustness both to non-editorial transformations \textit{and} imperceptible adversarial perturbations, preventing attacks illustrated in Fig.~\ref{fig:teaser}.  We show that this algorithm improves adversarial robustness of both tamper-sensitive \cite{nguyen2021oscar} and tamper-invariant \cite{black2021deep} image fingerprinting models, and we also discuss how to make the image comparator model \citep{black2021deep} robust.  The approach is conceptually simple and leads to a relatively small computational overhead ($\approx$ $2 \times$ slowdown) compared to standard contrastive learning.  We also show that our adversarially-robust image hashing models have benefits in terms of interpretability: they output perceptually similar images under hash inversion attacks \citep{strupek2021learning}. 
\end{enumerate}

Our work %
comprehensively studies adversarial robustness for the growing body of image attribution work, and is timely given the emergence of cross-industry standards reliant, in part, on such attribution.  For example, the Coalition for Content Provenance and Authenticity (C2PA) \cite{c2pa} proposes to fight misinformation by embedding secure audit information on content manipulation within image metadata.  Many online media distribution channels (such as social media sites)  strip away this metadata.  C2PA proposes to counter this by the use of perceptual hashes for image attribution, and advocates computing the hashes on the client side due to privacy concerns.  This implies, as with our work, that the attacker has \textit{white-box} access to the target model (\ie the attacker is assumed to know all details of the system being attacked).  Our work is the first to demonstrate the significance of these attacks on attribution models. Moreover, by offering the first defense against adversarial attacks for such  models, we contribute to the protection of provenance systems implementing such standards.

\section{Related Work}

\paragraph{Image fingerprinting for provenance.}
Image fingerprinting models robust to non-editorial transformations were proposed in \citet{black2021deep} and \citet{nguyen2021oscar}. These represent two complementary approaches to applying image retrieval to the  attribution problem. Both approaches match query images to a trusted database of originals, invariant to non-editorial changes such as resolution, format or quality change. However, the approaches differ in their consideration of manipulated images. \citet{black2021deep} train the image retrieval model to match manipulated images to originals successfully, i.e., to bring such images pairs close together in the search embedding.  By contrast, \citet{nguyen2021oscar} train the model to separate such image pairs, encouraging matching to fail in the presence of content manipulation. The advantage of \cite{nguyen2021oscar} is a simpler pipeline as it has only an image retrieval model (with an optional geometric verification step), while \cite{black2021deep} relies on a separate image comparator network that analyzes whether a pair of images are identical, different or have been manipulated, visualizing the difference. Whilst robust to non-editorial distortions, both approaches are not robust to adversarial attacks, motivating our work.

{\bf Adversarial attacks} on deep neural networks for image classification were pioneered by \citet{szegedy2013intriguing}, who demonstrated that minor perturbations of pixel values are sufficient to induce significant classification mistakes despite little perceptible difference (`covert' approaches).  \citet{Goodfellow2014} demonstrated linearity of this effect in input space, introducing the fast gradient sign method (FGSM) to quickly compute adversarial perturbations via backpropagation without the need for solving costly optimizations. An iterative form of this method for more robust attacks was later presented \cite{BasicIterative}.  Such attacks have received significant attention in recent years with many variants proposed to covertly attack image classifiers \cite{DeepFool,gowal2019alternative,croce2020reliable}.  Adversarial patches take a complementary, `overt', approach via synthesis of vivid  `stickers' \cite{AdversarialPatch} that occupy only a small  region yet induce misclassification \cite{AdversarialPatch,Eykholt2017} or misdetection \cite{chen2018robust,Thys2019}.

Recently, adversarial attacks on image retrieval models have been demonstrated via similar means. \citet{tolias2019targeted} show that image retrieval models are non-robust. They perform targeted attacks in the white-box and semi black-box setting (unknown pooling).  
\citet{bai2020targeted} and \citet{dolhansky2021adversarial} show that image \textit{hashing} models can be fooled as well, including attacks which exactly produce target hashes.

\paragraph{Contrastive learning.} Several popular self-supervised learning approaches are based on contrastive learning: SimCLR \citep{chen2020simple}, MoCo \citep{he2020momentum}, and BYOL \citep{grill2020bootstrap}.
Most robust self-supervised approaches focus on robust transfer learning \citep{Hendrycks2019Using, chen2020adversarial, kim2020adversarial, jiang2020robust, xu2020adversarial, ho2020contrastive, gowal2021selfsupervised} or multi-objective optimization \citep{hendrycks2019usingself, mao2019metric, bui2021understanding} to improve adversarial robustness.
The focus of these works differ from our focus on image retrieval. In particular, they do not benchmark image retrieval performance and for training, they rely on augmentations that are optimized for transfer learning and not for image attribution (e.g., extreme crops used in SimCLR: between 8\% and 100\% of the area as in \citet{chen2020simple}).
E.g., \citet{kim2020adversarial} propose a two-stage robust training approach: first generating instance-wise adversarial examples for the SimCLR loss and then combining together the standard SimCLR loss with the robust loss. 
\citet{tamkin2021viewmaker} propose to use an image-to-image network that generates adversarial examples which are then projected onto a (relatively large) $\l_1$-ball and they do not cover adversarial robustness.
However, neither of these approaches is considered in the image retrieval setup.
\citep{robinson2021can} propose  adversarial training in the \textit{latent} space with the goal of improving standard generalization.
Closer to image attribution, \citet{panum2021exploring} combine deep metric learning algorithms with adversarial training but they perform only small-scale experiments and evaluate their models via nearest neighbour classification which is different from the retrieval under editorial and non-editorial distortions as we do in the context of image attribution.

\begin{figure*}[t!]
    \centering
    \includegraphics[width=0.95\linewidth]{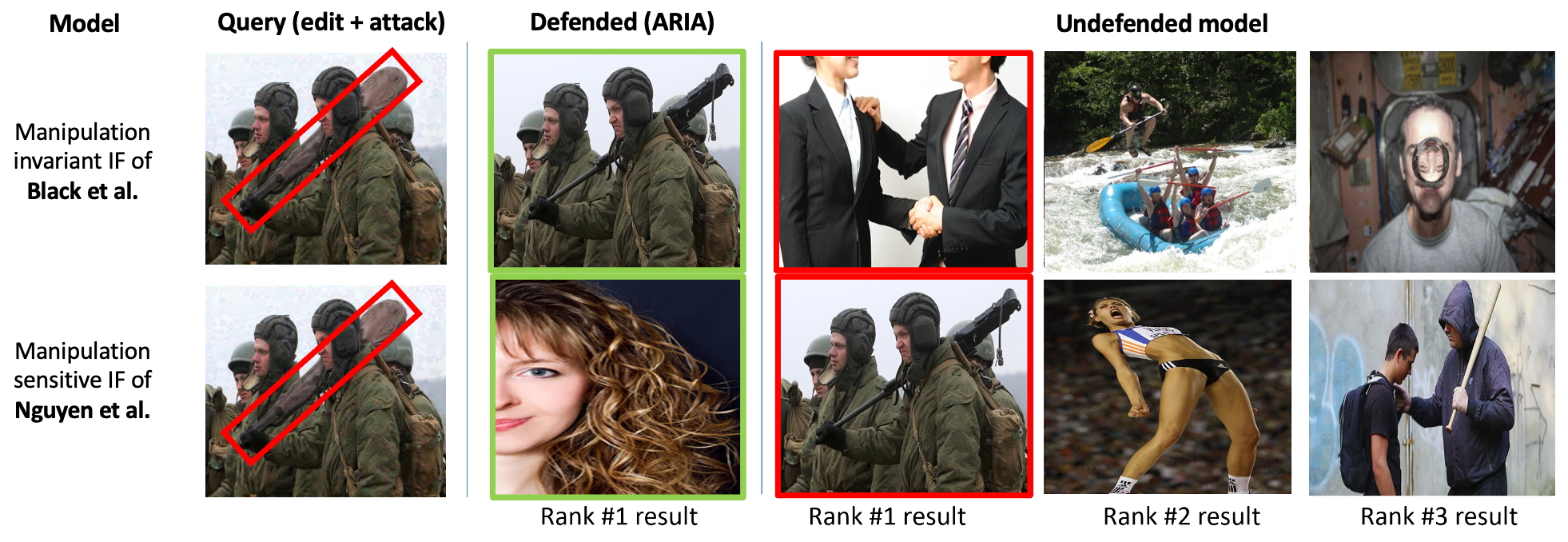}
    \caption{\textbf{Attacks on image fingerprinting (IF) models.} Visualization of untargeted adversarial examples of size $\eps_\infty=\nicefrac{8}{255}$ on two IF approaches with complementary goals. Upper: a model seeking to match an original image, invariant to any editorial change in the query \cite{black2021deep}. Lower: a model seeking to {\em avoid} matching an edited query to the original \cite{nguyen2021oscar}.  In both cases it is possible to attack the model to defeat the goal, and in both cases ARIA defends it successfully.}
    \label{fig:attack_if}
    \squeezeup
\end{figure*}
\begin{figure*}[t!]
    \centering
    \includegraphics[width=0.93\linewidth]{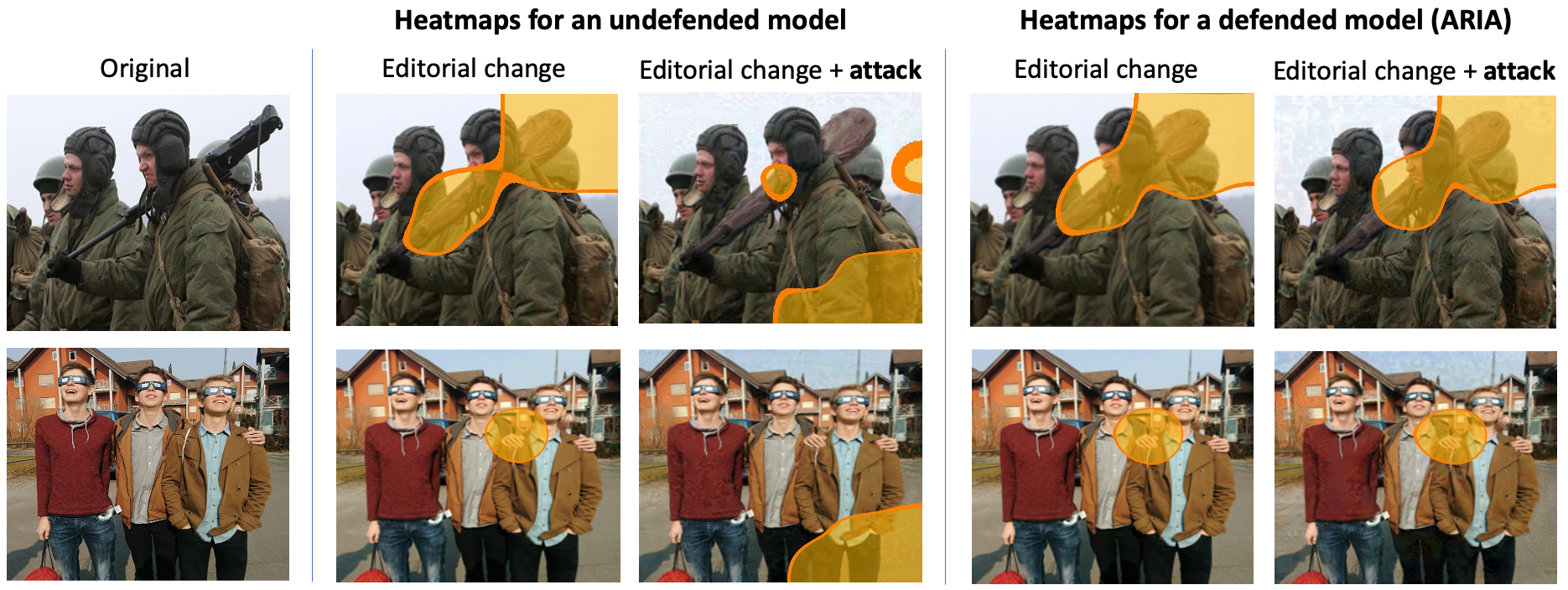}
    \caption{\textbf{Attack on the image comparator (IC) model of \citet{black2021deep}.} Visualization of the heatmap generated for an undefended and ARIA defended models queried using an adversarial example of budget $\eps_\infty=\nicefrac{8}{255}$. The attack targets a heatmap prediction within the bottom-right quadrant of the image. Shown for two different images drawn from the PSBattles dataset. In all cases the ARIA defended model predicts a heatmap near-identical to the original heatmap inferred by the undefended model in the absence of the attack. }
    \label{fig:attack_ic}
    \squeezeup
\end{figure*}
\section{Vulnerability of Attribution Models}

We start from studying adversarial vulnerability within the context of the image attribution approaches of \citet{nguyen2021oscar} and \citet{black2021deep}.

\textbf{1. Image fingerprinting (IF)}. We consider an IF model $\phi: \R^d \rightarrow \R^D$ which performs the mapping of an image to its $D$-dimensional feature vector (fingerprint) used to output the most similar image from a database $\mathcal{X}$. We denote by $x$ an original image contained in $\mathcal{X}$, and $x_{query}$ the image $x$ modified by some transformation.  For all IF methods, we wish the match to be invariant to a set of non-editorial transformations $\bar{E}(x)$, 
and in some cases (\eg \citet{black2021deep}) also editorial manipulations of the image ${E}(x)$.  

In all cases we consider adversarial perturbations of $x$ specific to the model $\phi$ generated under some perturbation budget $\epsilon$. We call the perturbations \textit{targeted} or \textit{untargeted} perturbations depending on whether an attack targets retrieval of a specific incorrect image, or its objective is simply to prevent retrieval of the correct image. In the case of methods seeking to retrieve $x$ for $x_{query} \in \bar{E}(x)$ only \citep{nguyen2021oscar}, a specific form of targeted attack attempts to fool $\phi$ to retrieve its original as illustrated in Fig.~\ref{fig:attack_if}.

\textbf{2. Image comparison (IC).} IC methods visualize pixel regions containing \textit{editorial} changes, performing an `intelligent differencing' operation between $x_{query}$ and the top retrieval from the IF model, ignoring any visual change due to non-editorial transformations. We consider the approach of \cite{black2021deep} where the model outputs both such a visualization (a $7\times7$ heatmap) 
and additionally assigns the image pair to three categories: same images with non-editorial changes, same images with editorial changes or different images.  The first goal of the attacker is to make the comparator classify an image with editorial changes to the first category. The second goal is to make the comparator output a misleading heatmap describing editorial changes (see Fig.~\ref{fig:attack_ic}).

\subsection{Adversarial Attack Scope}

We consider attacks within the following scope. First, adversarial perturbations should be \textbf{imperceptible} such as perturbations bounded within a small $\l_\infty$- or $\l_2$-norm. Imperceptibility is a crucial property as, from an attack perspective, a user should not realize an image has been manipulated in any way. This requirement makes, e.g., patch-based \citep{AdversarialPatch}, $\ell_1$- or $\ell_0$-bounded perturbations not relevant in our case since they render tampering visually obvious. Second, we consider a \textbf{white-box} knowledge model: all details of the model are assumed known. This prevents the so-called \textit{security-by-obscurity} where the security or robustness of a system relies on such detail (e.g., model architecture or parameters) being secret. In practice, such detail can be leaked or reverse engineered, particularly for models deployed on edge devices or client-side (as is advocated in emerging standards \cite{c2pa}). Third, we focus on \textbf{fully realizable attacks}, i.e., after generating adversarial examples, we save them as valid JPEG files and only then evaluate the model on them.  This differs from most prior works, where imperceptible adversarial images are generated such that they have arbitrary real values in the range $[0, 1]$ \citep{szegedy2013intriguing, madry2018towards}. Instead, after saving them as JPEG files, the pixel values are quantized to 8-bit and the image is compressed introducing further non-editorial changes. 
We show in Sec.~\ref{sec:exp_setup} that even standard attacks described below (as opposed to robust attacks \citep{athalye2018synthesizing}) are sufficient for the considered attack scenario as we can successfully reduce many performance metrics to zero. %

\subsection{Implementation of adversarial attacks}
We generate adversarial attacks using projected gradient descent (PGD) \citep{madry2018towards} under $\l_\infty$-norm constraints since the gradients are available\footnote{We use differentiable image resizing to make sure that the whole image preprocessing pipeline is differentiable.} in the white-box setting and the $\ell_\infty$-norm is a useful proxy for the imperceptibility requirement.

\textbf{Image fingerprinting attack.} The goal of an untargeted attack is to make the resulting adversarial example $x_{query}+\delta$ have an IF sufficiently different from the original one so that $x_{query}+\delta$ gets matched to an incorrect image. For this, we can \textit{maximize} the $\l_2$-distance between an IF of $x_{query}+\delta$ and the IF of the original $x$: 
\begin{eqnarray}
    \text{max \ \ } &\norm{\phi(x_{query}+\delta) - \phi(x)}_2^2 \\ \nonumber
    \text{s.t. \ \ } &\norm{\delta}_\infty \leq \eps, \ \ \ 0 \leq x_{query} + \delta \leq 1.
\end{eqnarray}
In Fig.~\ref{fig:attack_if} (upper), we show results of such an untargeted attack on \citet{black2021deep}. The method attempts to match query images exhibiting both editorial and non-editorial change, back to an original. The attack successfully defeats this goal.  

By contrast, in Fig.~\ref{fig:attack_if} (lower)  the model of \citet{nguyen2021oscar}  aims to avoid matching an edited image to an original, in order to avoid corroborating a manipulated image with the provenance of its original. We perform a \textit{targeted} attack which attempts to match a query image $x_{query}$ to the original image $x$:
\begin{eqnarray}
    \label{eq:objective_for_oscarnet}
    \text{min \ \ } &\norm{\phi(x_{query}+\delta) - \phi(x)}_2^2 \\ \nonumber
    \text{s.t. \ \ } &\norm{\delta}_\infty \leq \eps, \ \ \ 0 \leq x_{query} + \delta \leq 1.
\end{eqnarray}
We note here that when producing adversarial examples for the OSCAR-Net of \citep{nguyen2021oscar}, we assume the object detector's output to be fixed (see the details in the sup. mat).

\textbf{Image comparator attack.}
To attack the prediction module of an IC model \citep{black2021deep} (Fig.~\ref{fig:attack_ic}), we minimize the probability $p_C$ of the ground truth class $y$ (e.g., `same images with editorial changes') over the perturbation $\delta$ added only to $x_{query}$: %
\begin{eqnarray}
    \text{min \ \ } & \log p_C(x_{top-1}, x_{query}+\delta)_y \\ \nonumber
    \text{s.t. \ \ } &\norm{\delta}_\infty \leq \eps, \ \ \ 0 \leq x_{query} + \delta \leq 1.
\end{eqnarray}
For the heatmap attack, we minimize the cosine similarity between the predicted heatmap $f_T(x)$ and ground truth $t \in [0, 1]^{7\times7}$ since this is the loss used for training in \citep{black2021deep}:
\begin{eqnarray}
    \text{min \ \ } & \cos(f_T(x_{query}+\delta), t) \\ \nonumber
    \text{s.t. \ \ } &\norm{\delta}_\infty \leq \eps, \ \ \ 0 \leq x_{query} + \delta \leq 1.
\end{eqnarray}
Moreover, targeted attacks on heatmaps may trick a user into distrusting an \textit{original} image.
To generate targeted adversarial examples, we can just flip the minimization to maximization and use some target label or heatmap.

\section{Adversarially Robust Image Attribution}\label{sec:detailed_description}
In this section, we propose a robust contrastive learning algorithm  applicable to both fingerprinting approaches described in  \citet{nguyen2021oscar} and \citet{black2021deep}, including the image comparator of the latter.

\subsection{Robust contrastive learning}
We propose a method that adapts contrastive learning with the SimCLR loss \citep{chen2020simple} to be robust to imperceptible adversarial examples.
Denote by $L(\{x_i\}_{i=1}^{2N})$ the SimCLR loss defined on a batch of paired \textit{positive} examples, where $i$-th and $(N+i)$-th examples correspond to the same images but with different \textit{transformations}:
\begin{align}
    \nonumber
    &L(\{x_i\}_{i=1}^{2N}) = \frac{1}{2N} \sum_{i=1}^N [ \l(\{x_i\}_{i=1}^{2N})_{i,N+i}  +\l(\{x_i\}_{i=1}^{2N})_{N+i,i}], \\ 
    &\l(\{x_i\}_{i=1}^{2N})_{i,j} = -\log\frac{e^{\text{cos}(\phi(x_i), \phi(x_j))/\tau}}{\sum_{k=1}^{2N} \mathbbm{1}_{k \neq i} e^{\text{cos}(\phi(x_i), \phi(x_k))/\tau}}
\end{align}
These transformations are of two kinds: non-editorial $\bar{E}(x)$ (e.g., affine transformations and ImageNet-C \citep{hendrycks2019robustness}) and editorial $E(x)$ (e.g., available as paired images in the PS-Battles dataset \citep{heller2018ps}).
\citet{black2021deep} use both of them as \textit{positive} examples while \citet{nguyen2021oscar} only treats images with non-editorial changes as positive.

Then to train adversarially robust IF models, we change the objective following the robust optimization framework \citep{madry2018towards} by adding an inner loop to maximize the loss on adversarially perturbed images $x_i + \delta_i$: 
\begin{align}
    &\min_{\theta \in \R^{|\theta|}} \E_{\{x_i\}_{i=1}^{2N} \sim D} \Big[ \max_{\substack{\norm{\delta_i}_\infty \leq \eps \\ 0\leq x_i+\delta_i \leq 1}} L(\{x_i+\delta_i\}_{i=1}^{2N}) \Big],
\end{align}
where $\theta$ denotes the model parameters and $D$ the data distribution.
We note that although adversarial training \citep{madry2018towards} is an established technique for image classification, IF models are trained differently. 
Thus, it is not clear in advance if the findings and pitfalls of robust training of image classification models (e.g., such as catastrophic or robust overfitting \citep{Wong2020Fast, Rice2020Overfitting}) transfer to the IF setting.

To solve the inner maximization problem, we use a few iterations of projected gradient ascent (in practice, up to 3) for the inner maximization problem, where each iteration requires an evaluation of the input gradient $\nabla_{\delta_i} L(\{x_i + \delta_i\}_{i=1}^{2N})$ via backpropagation. Using a few iterations of the attack comes out to be sufficient to prevent the \textit{catastrophic overfitting} problem \citep{Wong2020Fast, Andriushchenko2020Understanding} which also manifests itself in training IF models, as we observe in the experimental part. %

The final objective that we use combines the robust version of the SimCLR loss \cite{chen2020simple} with the hashing term from \citep{nguyen2021oscar} for large-scale search has the following form:
\begin{align}
    \label{eq:simclr_final_obj}
    \min_{\theta \in \R^{|\theta|}} &\E_{\{x_i\}_{i=1}^{2N} \sim D} \Big[ \max_{\substack{\norm{\delta_i}_\infty \leq \eps \\ 0\leq x_i+\delta_i \leq 1}} L(\{x_i+\delta_i\}_{i=1}^{2N}) \Big] + \\ \nonumber
                                    &\alpha \E_{x \sim D} \left[ \norm{\phi(x) - \sign(\phi(x))}^3 \right].
\end{align}
For \cite{black2021deep}, which proposes no end-to-end hashing, we mostly report the models trained without the hashing term.
In practice, we approximate the expectations using mini-batches, and we apply the hashing term on the same examples as the main loss.
We do not use projection layers on top of the target embeddings as in \citet{chen2020simple} since we found this leads to worse performance.

\subsection{Robust image comparator network}
Next we discuss how to make the \textit{image comparator model} from \citet{black2021deep} robust to adversarial attacks. First, we note that the image comparator performs a classification task (both for the prediction and heatmap modules) for which there are well-described solutions in the literature on how to improve their robustness \citep{madry2018towards, Zhang2019Theoretically, Gowal2020Uncovering}.
Thus, we use the multi-task objective of \citet{black2021deep} with the classification loss $w_c \mathcal{L}_C$ and heatmap loss $w_t \mathcal{L}_T$ (we use the same losses and weights as in \citet{black2021deep}, i.e. the cross-entropy and cosine similarity and $w_c=w_t=0.5$), but we add an inner maximization operator to it:
\begin{align}
    \nonumber
    \min_{\theta \in \R^{|\theta|}} \E_{x_1, x_2, y, t} \Big[ \max_{\substack{\norm{\delta}_\infty \leq \eps \\ 0\leq x_2+\delta \leq 1}} 
    &w_c \mathcal{L}_C (x_1, x_2+\delta, y) + \\
    &w_t \mathcal{L}_T (x_1, x_2+\delta, t)\Big],
\end{align}
where $y$ is the class label, $t$ is the ground-truth heatmap.
Note that we add adversarial perturbations only to the corrupted query image $x_2$ as we assume that the image comparator operates on the images $x_1$ from the database which contains non-adversarial original images. 
As corrupted images, we use either original images under a non-editorial distortion (see the experimental setup below), manipulated images from PSBattles or simply a different image under a non-editorial distortion.
We use SGD for the outer loop and a few steps of PGD to approximate the maximization. We note that the image comparator model is fully differentiable, including the geometric alignment RAFT module \citep{teed2020raft} which is a part of their model. Thus, we both train \textit{and} generate adversarial examples completely end-to-end for this network.

\section{Experimental evaluation}
\label{sec:exp_setup}
In this section, we provide a detailed description of the experimental setup, describe the main results and show ablation studies that give more insights in the proposed method.

\subsection{Experimental setup}
\paragraph{Performance metrics.}  
For the image retrieval model of \citet{black2021deep}, we similarly consider recall-based metrics assuming that for all queried images, the originals  have been indexed:
(1) \textit{standard recall}: $\text{Pr}\left[f(x_{query}) = f(x)\right]$ which is the probability of retrieving a correct image under some image corruption (non-editorial transformation, manipulation or both),
(2) \textit{adversarial recall}: $\text{Pr}\left[f(x_{query}) + \delta = f(x)\right]$ which is the probability of retrieving a correct image under non-editorial transformations and an adversarial perturbation $\delta$.
For the image comparator model \citep{black2021deep}, we use the same metrics as in their paper: \textit{average precision} (AP) for the classification module and the \textit{intersection over union} (IoU) over the images with editorial changes for the heatmap module. We use all images for the former and only images with editorial manipulations for the latter.

We note that unlike \citep{black2021deep}, OSCAR-Net \citep{nguyen2021oscar} %
is trained to distinguish non-editorial transformations from editorial manipulations.
Thus, we follow their metrics using standard mAP and top-1 recall (R@1) for the non-editorially transformed query set, also inverse mAP (imAP) and inverse recall (iR@1) for the tamper query set. For all metrics, the higher is better.

\paragraph{Training details.}
For training the robust image retrieval models we use the ResNet-50 architecture and Behance1M; a subset of 1M images sourced from a public digital art portfolio website (\texttt{behance.net}).\footnote{We will publish the URL list of these images upon acceptance.}  We use Behance1M for training since it is significantly larger than PSBattles \cite{heller2018ps} and more diverse, containing not only photos but also graphics. As a starting set of parameters, we use the model from \citet{black2021deep} pre-trained for 2 epochs with standard contrastive training.
We train the model to be robust not only to adversarial examples (after two epochs of standard training) but also to \textit{non-editorial transforms} via data augmentation, for which we use the \texttt{beacon\_aug} library \citep{beacon-aug2021} to apply ImageNet-C corruptions \citep{hendrycks2019robustness} then resizing, rotation, padding, cropping, horizontal flips, and JPEG compression.

\paragraph{Evaluation details.}
We evaluate robustness on Behance and on PSBattles \citep{heller2018ps}; we use the `hard' subset of the latter defined in \citet{black2021deep}.
As in past work \cite{black2021deep, nguyen2021oscar}, we add distractor images from stock photography (Adobe Stock thumbnails).
We use the FAISS library \citep{johnson2017billion} for efficient image retrieval.
We generate adversarial attacks using $\l_\infty$ PGD with 50 iterations using $\eps=\nicefrac{8}{255}$, step size $\nicefrac{4}{255}$ decayed by a factor of 2 at 25\%, 50\%, and 75\% of iterations. 
Unless specified otherwise, we evaluate the image retrieval models on full PSBattles with 2M and 100K Adobe Stock distractors, following the settings in \cite{black2021deep} and \cite{nguyen2021oscar}, respectively.
We apply adversarial attacks in the original pixel space with differentiable resizing to $224\times224$ and JPEG-90 compression after it (thus, it is a \textit{fully realizable} attack). %

\begin{table*}[t!]
    \centering
    \footnotesize
    \setlength\tabcolsep{2.5pt}

    \begin{tabular}{l|cccc|cccc|cccc}
         \multicolumn{1}{c}{} & \multicolumn{12}{c}{\textbf{Top-1 and top-100 recall for different query sets}} \\
         & \multicolumn{4}{c|}{\textbf{Non-editorial distortions}} & \multicolumn{4}{c|}{\textbf{Editorial manipulations}} & \multicolumn{4}{c}{\textbf{Editorial + non-editorial}} \\
                                  & \multicolumn{2}{c}{\textbf{No attack}} & \multicolumn{2}{c|}{\textbf{$\l_\infty$ adversarial}}  &  \multicolumn{2}{c}{\textbf{No attack}} & \multicolumn{2}{c|}{\textbf{$\l_\infty$ adversarial}}  &  \multicolumn{2}{c}{\textbf{No attack}} & \multicolumn{2}{c}{\textbf{$\l_\infty$ adversarial}} \\
        \textbf{Existing models}  & \textbf{R@1} & \textbf{R@100} & \textbf{R@1} & \textbf{R@100}  &  \textbf{R@1} & \textbf{R@100} & \textbf{R@1} & \textbf{R@100}  &  \textbf{R@1} & \textbf{R@100} & \textbf{R@1} & \textbf{R@100}  \\
        \hline
        Standard supervised, ImageNet \citep{paszke2017automatic}  &  20.8 & 39.6 & 0.0 & 0.1  &  87.2 & \textbf{95.4} & 0.0 & 0.2  &  15.1 & 33.2 & 0.0 & 0.1 \\
        DeepAugment + AugMix supervised, ImageNet \citep{Hendrycks2020Many}  &  47.9 & 66.9 & 0.1 & 0.3  &  86.7 & 95.3 & 0.0 & 0.1  &  36.5 & 58.0 & 0.0 & 0.3 \\
        Robust supervised, $\eps_\infty=\nicefrac{4}{255}$, ImageNet \citep{Salman2020Do}  &  39.4 & 53.0 & 10.2 & 20.8  &  86.3 & 95.0 & 31.2 & 56.1  &  30.8 & 46.5 & 5.5 & 15.2 \\
        Undefended contrastive, PSBattles \citep{black2021deep}  &  74.1 & 93.7 & 0.0 & 0.0  &  80.1 & 92.9 & 0.0 & 0.0  &  55.3 & 83.5 & 0.0 & 0.0 \\
        
        \textbf{Our new models} \\
        \hline
        Undefended contrastive, Behance1M (ours)  &  \textbf{97.8} & \textbf{98.8} & 1.3 & 12.4  &  88.6 & 92.4 & 0.4 & 4.5  &  84.7 & 89.7 & 1.0 & 9.5 \\
        ARIA contrastive + hashing, $\eps_\infty=\nicefrac{4}{255}$, Behance1M  &  93.5 & 96.2 & 74.7 & 81.9  &  87.5 & 92.8 & 76.4 & 86.5  &  79.2 & 87.6 & 56.3 & 70.7 \\
        ARIA contrastive + hashing, $\eps_\infty=\nicefrac{8}{255}$, Behance1M  &  89.0 & 93.5 & 80.7 & 83.4  &  87.0 & 92.5 & 80.6 & 87.9  &  74.8 & 84.7 & 57.9 & 72.5 \\
        ARIA contrastive, $\eps_\infty=\nicefrac{2}{255}$, Behance1M  &  97.3 & 98.2 & 79.0 & 82.1  &  90.8 & 93.6 & 81.0 & 86.4  &  \textbf{87.3} & \textbf{91.0} & 63.0 & 71.0 \\
        ARIA contrastive, $\eps_\infty=\nicefrac{4}{255}$, Behance1M  &  96.4 & 97.3 & 83.0 & 85.7  &  \textbf{91.6} & 93.9 & 85.1 & \textbf{89.9}  &  86.7 & 90.3 & \textbf{69.7} & \textbf{77.0} \\
        ARIA contrastive, $\eps_\infty=\nicefrac{8}{255}$, Behance1M  &  94.2 & 96.0 & \textbf{83.7} & \textbf{87.4}  &  90.4 & 93.3 & \textbf{85.5} & \textbf{89.9}  &  83.1 & 88.0 & 69.3 & \textbf{77.0}
    \end{tabular}
    \caption{Standard and $\l_\infty$ adversarial ($\eps_\infty=\nicefrac{8}{255}$) top-1 and top-100 recall for different ResNet-50 models evaluated on PSBattles \citep{heller2018ps}. The database contains original images from PSBattles and 2M distractor images from Stock indexed using the IVF1024, PQ16 index from FAISS library following \citet{black2021deep}. We use three query sets based on PSBattles: (1) non-editorial distortions (ImageNet-C and affine) on original images, (2)  editorial manipulations but no distortions, (3)  editorial manipulations with non-editorial distortions.}
    \label{tab:recall_cvprw_models_approx_index}
\end{table*}
\subsection{Robust retrieval: \citet{black2021deep} approach}
\label{sec:cvprw_ret_exp}
\paragraph{Large-scale robustness evaluation.}
The main evaluation results for the robust image retrieval models trained on Behance1M are presented in Table~\ref{tab:recall_cvprw_models_approx_index} where we measure recall with 2M distractor images using the IVF1024, PQ16 index from FAISS library following \citet{black2021deep} (we include non-FAISS results also in the sup.mat., omitted as too slow to be practical). 
We report models trained with different $\eps_\infty$ since we want to show models with a different \textit{robustness-accuracy tradeoff} \citep{tsipras2019robustness}.
We measure adversarial recall under perturbations of size $\eps_\infty=\nicefrac{8}{255}$ since this is the most commonly used perturbation size in the literature \citep{madry2018towards, croce2021robustbench}.
Compared to the existing IF models trained contrastively on PSBattles \citep{black2021deep} and in a supervised way on ImageNet \citep{paszke2017automatic, Hendrycks2020Many, Salman2020Do}, 
our proposed method allows us to train IF models which are robust to imperceptible adversarial perturbations \textit{in addition} to being highly accurate.
For example, our robust model trained with $\eps_\infty=\nicefrac{4}{255}$ achieves 91.6\% standard and 85.1\% adversarial recall compared to 80.1\% and 0.0\% achieved by the main baseline from \citep{black2021deep}.
In addition, Table~\ref{tab:recall_cvprw_models_approx_index} shows that robust contrastive learning with $\eps=\nicefrac{4}{255}$ leads to a higher recall for manipulated images compared to the standard model even despite having a worse recall on non-editorial distortions. We note that this shows another interesting case when adversarial training can improve the performance on \textit{real-world distribution shifts} in addition to the benefits of adversarial training known before on, e.g., common corruptions \citep{ford2019advnoise, xie2020adversarial, Kireev2021Effectiveness} or transfer learning \citep{Salman2020Do, utrera2020adversarially}.
Finally, the models trained with the hashing term (see Eq.~\eqref{eq:simclr_final_obj}) perform slightly worse than the models producing real-valued IFs, but there can still be interesting use-cases for them such as faster search, lower memory requirements, and a potential storage in key-value data structures. %

\begin{table}[t]
    \centering
    \footnotesize
    \setlength{\tabcolsep}{4pt}
    \begin{tabular}{l c c c c c}
     & & \multicolumn{4}{c}{\textbf{Adversarial recall}} \\    
     & & $\eps_2$ & $\eps_\infty$ & $\eps_\infty$ \\ 
     \textbf{Models} & \textbf{Recall} & $5.0$ & $\nicefrac{16}{255}$ & $\nicefrac{32}{255}$ \\
     \hline
     Undefended, PSBattles \citep{black2021deep} & 92.2\% & 0.0\% & 0.0\% & 0.0\% \\
     ARIA, $\eps_\infty=\nicefrac{2}{255}$, Behance1M & \textbf{99.6\%} & \textbf{84.2\%} & 38.8\% & 2.8\% \\
     ARIA, $\eps_\infty=\nicefrac{4}{255}$, Behance1M & 98.4\% & 84.0\% & 50.6\% & 9.8\% \\
     ARIA, $\eps_\infty=\nicefrac{8}{255}$, Behance1M & 97.8\% & 83.4\% & \textbf{60.0\%} & \textbf{15.4\%} \\
    \end{tabular}
    \caption{Standard and adversarial top-1 recall for attacks \textit{unseen} during training. We use evaluation on a query set of non-editorial distortions of Behance1M images (500 distractors).
    }
    \label{tab:unseen_attacks}
    
\end{table}
\paragraph{Robustness to unseen adversarial perturbation.}
In Table~\ref{tab:unseen_attacks}, we show the robustness results for perturbations which were unseen during training such as $\l_2$-bounded perturbations ($\eps_2 = 0.5$) and $\l_\infty$-perturbations of a larger radius compared to those used for training ($\eps_\infty \in \{\nicefrac{12}{255}, \nicefrac{16}{255}, \nicefrac{32}{255}\}$).
We can see that robustness indeed generalizes to these other types of perturbations: e.g., $\eps_2 = 0.5$ is sufficient to reduce the adversarial recall to $0.0\%$ for the undefended model of \citet{black2021deep} while all our $\ell_\infty$-trained robust models achieve $83\%+$ adversarial recall.

\begin{figure}[t!]
    \centering
    \scriptsize
    \begin{subfigure}[c]{.32\linewidth}
        \caption{\scriptsize \textbf{Original image $x$}}
        \includegraphics[width=\textwidth]{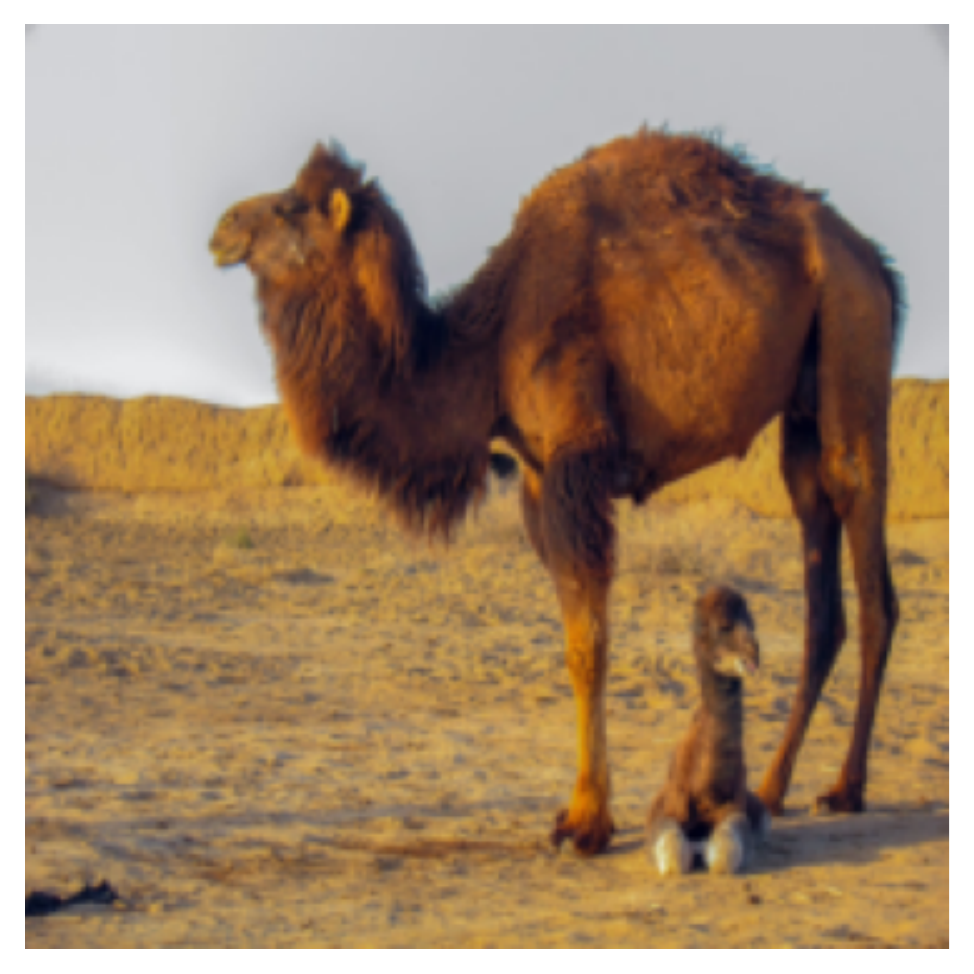}
        \includegraphics[width=\textwidth]{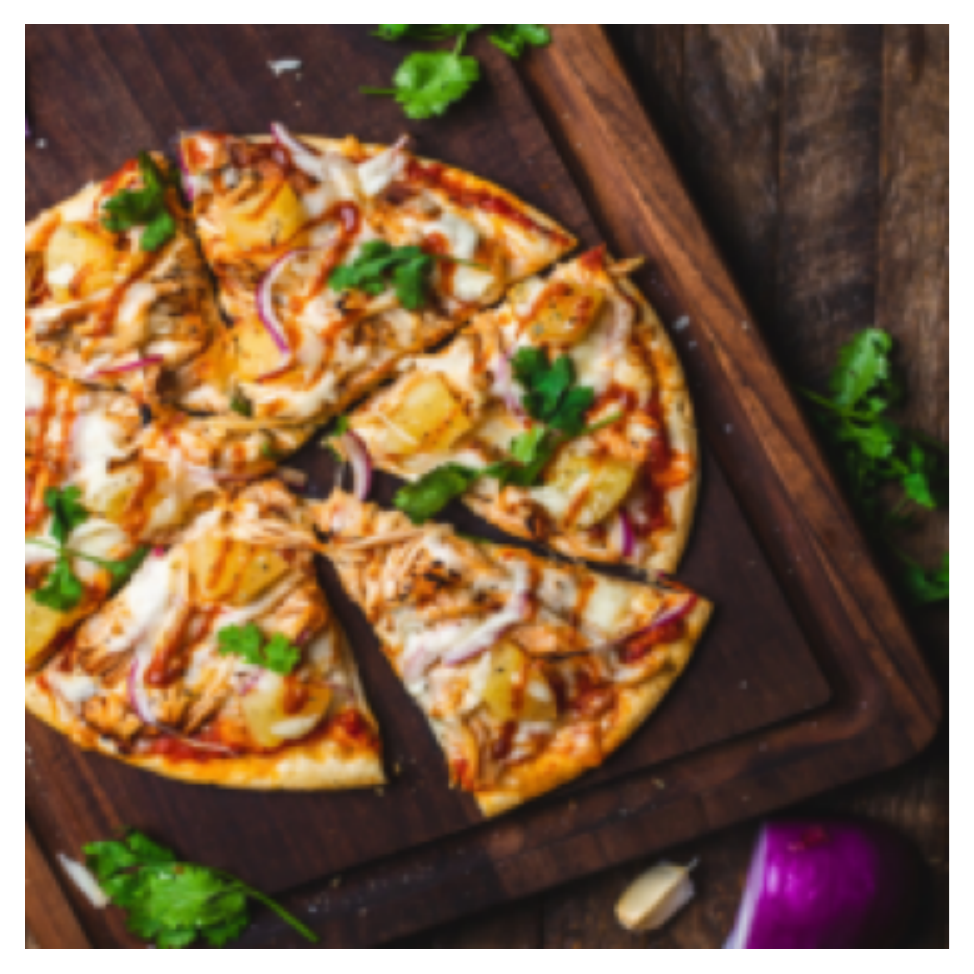}
    \end{subfigure}
    \begin{subfigure}[c]{.32\linewidth}
        \caption{\scriptsize \textbf{$\phi^{-1}_{\approx}(x)$, standard model}}
        \includegraphics[width=\textwidth]{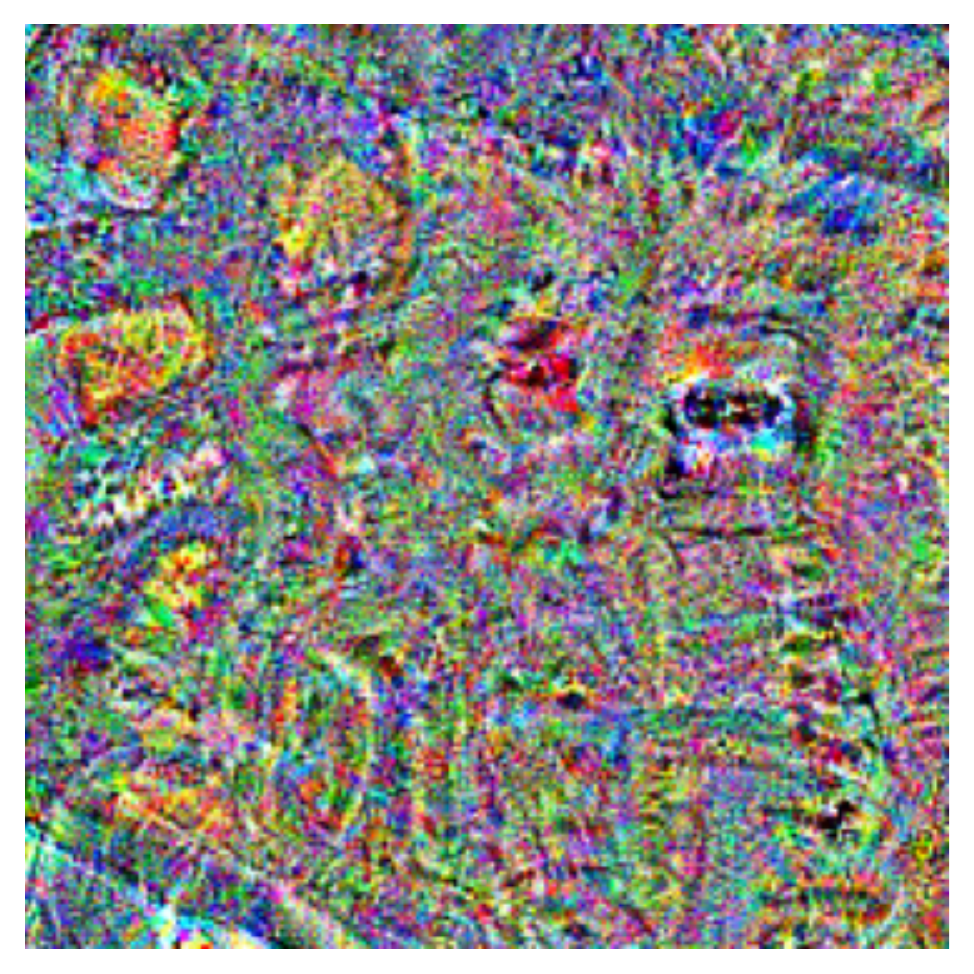}
        \includegraphics[width=\textwidth]{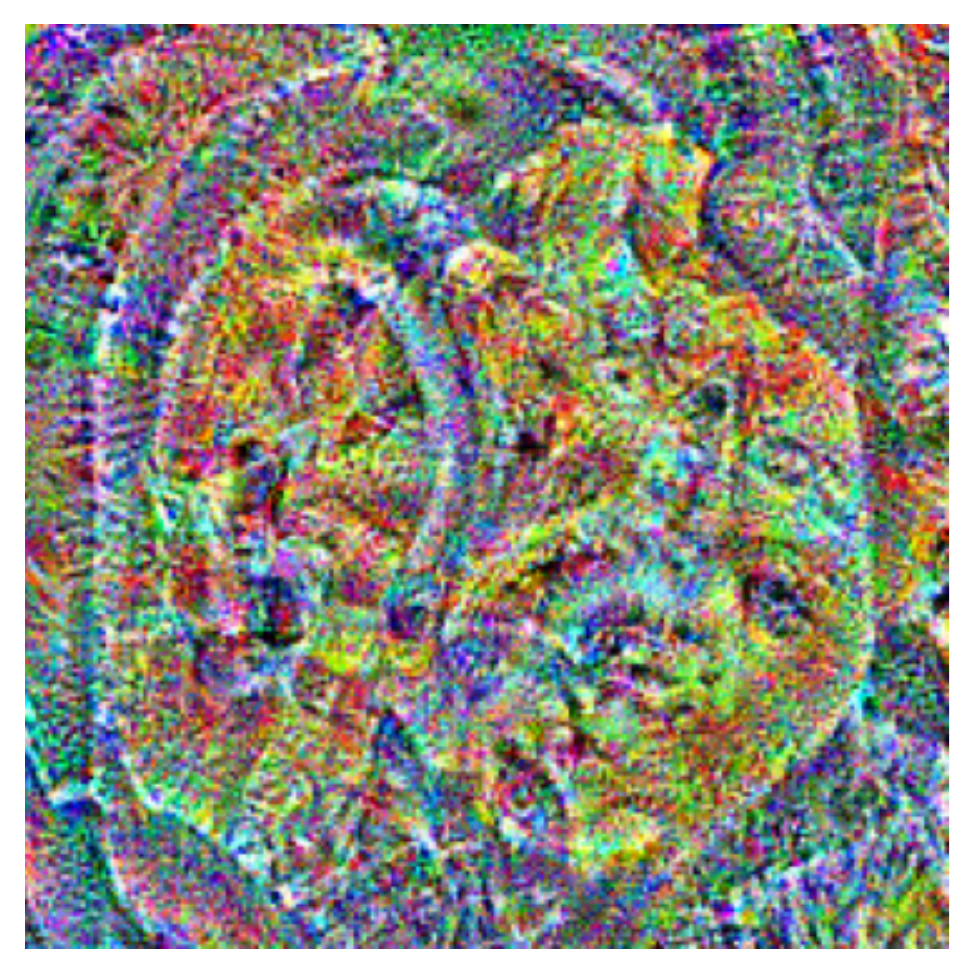}
    \end{subfigure}
    \begin{subfigure}[c]{.32\linewidth}
        \caption{\scriptsize \textbf{$\phi^{-1}_{\approx}(x)$, robust model}}
        \includegraphics[width=\textwidth]{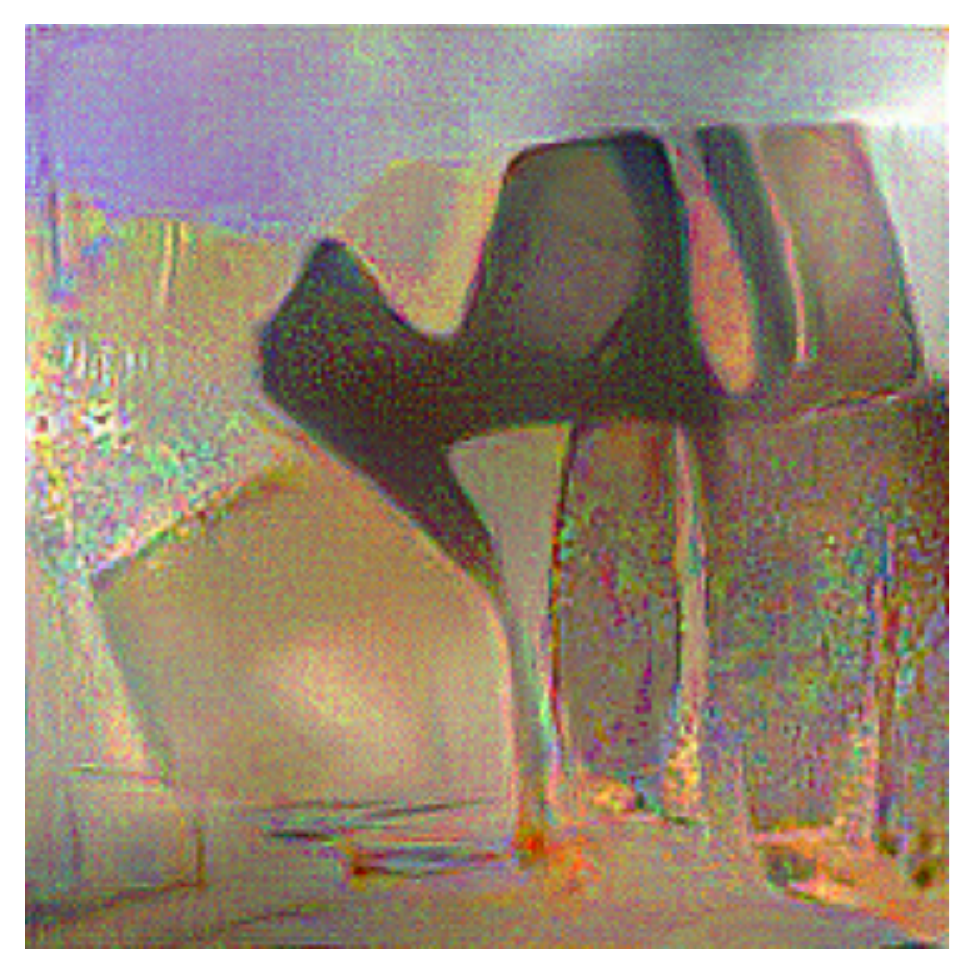}
        \includegraphics[width=\textwidth]{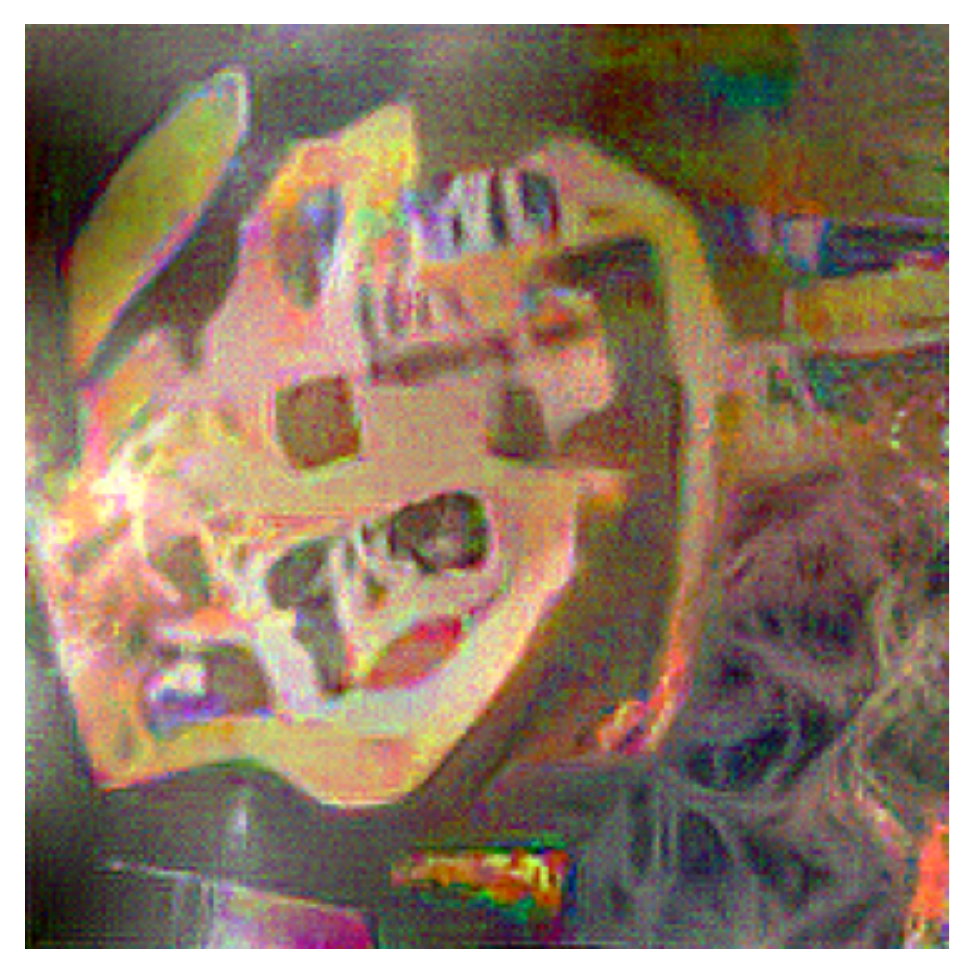}
    \end{subfigure}
    \centering
    \caption{Visualization of the hash inversions $\phi^{-1}_{\approx}(x)$ (see Eq.~(\ref{eq:hash_inversion})) for two original images $x$ (\textbf{left}) for a standard model (\textbf{middle}) and ARIA model with $\eps_\infty=\nicefrac{4}{255}$ (\textbf{right}), both trained on Behance1M.}
    \label{fig:hashing_attack}
        \squeezeup
    \squeezeup
\squeezeup
\end{figure}
\paragraph{Plausible hash inversions.}
Here we show that adversarially robust image \textit{hashing} models output plausible images under the \textit{hash inversions attacks}. 
This type of attacks has recently received a lot of attention as a significant weakness of neural hashing models \citep{strupek2021learning}.
The goal of this attack is to compromise the validity of the hashing model by finding some \textit{irrelevant} images that lead to the \textit{same} binary hash as some target image $x$.
The formulation is similar to that of targeted adversarial attacks but without the $\l_\infty$-norm constraint on the perturbation magnitude and starting from some arbitrary $\hat{x}$ (we use a constant gray image). We take the robust model from Table~\ref{tab:recall_cvprw_models_approx_index} trained to output binary hashes via the sign function for which we use a differentiable approximation (\texttt{tanh} function with a parameter $\beta$):
\begin{align}
    \label{eq:hash_inversion}
    \phi^{-1}_{\approx}(x) = \argminop_{0 \leq \hat{x} \leq 1} \norm{\tanh(\beta \cdot \phi(\hat{x})) - \sign(\phi(x))}_2^2.
\end{align}
We solve the formulation using 1'000 iterations of PGD ensuring the exact hash match in each case.
We note that similar formulations in the context of image classification have been studied in \citep{mahendran2015understanding, dosovitskiy2016inverting, engstrom2019adversarial} from the interpretability perspective where they focused on inverting real-valued embeddings of a deep network instead of hashes.

As we can see from Fig.~\ref{fig:hashing_attack}, hash inversion attacks on standardly trained hashing models tend to produce obscure high-frequency patterns. %
At the same time, our robust image hashing models tend to focus more on shapes of objects which are approximately recovered under hash inversions. 
This behaviour is closely related to the adversarial vulnerability problem: the attacker can use non-robust features \citep{ilyas2019adversarial} to arbitrarily manipulate the model's hash. However, once we fix this problem via robust training, hash inversions start to be more related to the original images. %

\begin{figure}[t]
    \centering
    \includegraphics[width=0.85\columnwidth]{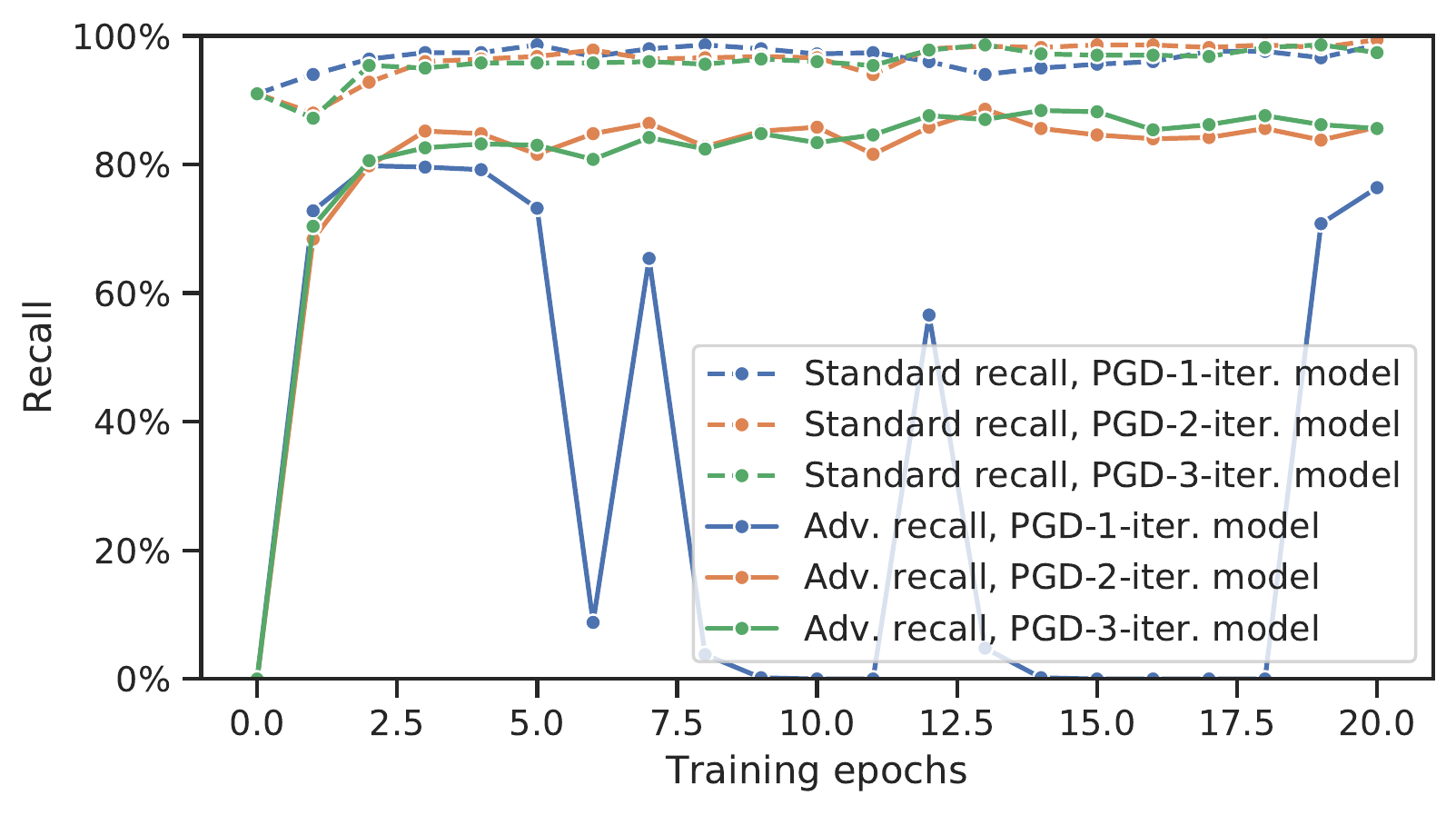}
    \includegraphics[width=0.85\columnwidth]{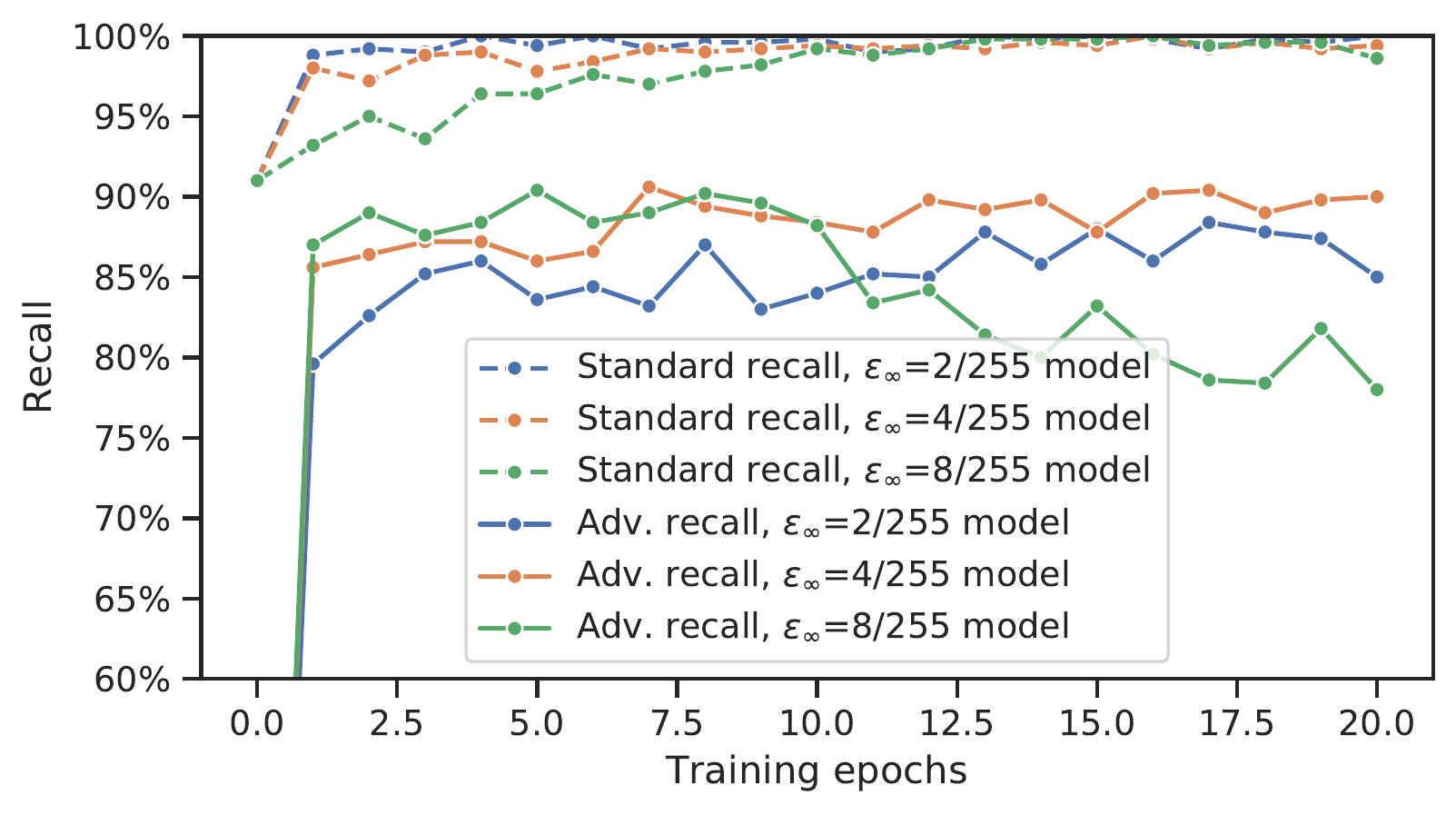}
    \caption{Standard and $\l_\infty$ adversarial ($\eps_\infty=\nicefrac{8}{255}$) top-1 recall at different epochs for models trained with (1) different numbers of PGD iterations (using $\eps_\infty=\nicefrac{4}{255}$) and (2) different perturbation radii $\eps_\infty$ (using 3 iterations of PGD) . We evaluate a query set of non-editorial distortions of Behance1M images (500 distractors).}
    \label{fig:if_hyperparameter_importance}
        \squeezeup
    \squeezeup
\end{figure}
\paragraph{Hyperparameter importance.}
Here we analyze the hyperparameters of ARIA training: the number of PGD iterations and the perturbation radius $\eps_\infty$. 
We train multiple models on Behance1M and report their baseline (\ie no attack) and adversarial recall (\ie for an attack with budget $\eps_\infty=\nicefrac{8}{255}$). 
Fig.~\ref{fig:if_hyperparameter_importance} (top) suggests that, similar to image classification, catastrophic overfitting \citep{Wong2020Fast, Andriushchenko2020Understanding} leads to unstable performance for robust contrastive learning, so training with a single iteration of PGD should be avoided.
We observe 2-3 iterations of PGD to be sufficient: which leads to the slowdown factor of $1.9\times$ and $2.3\times$, respectively. 
Fig.~\ref{fig:if_hyperparameter_importance} (bottom) shows that the model trained with $\eps=\nicefrac{8}{255}$ starts to \textit{overfit} in terms of adversarial recall after 7 epochs. This suggests that training with a too large $\eps_\infty$ may be problematic.

\begin{table*}[t!]
    \centering
    \footnotesize
    \setlength\tabcolsep{2.5pt}
    
    \begin{tabular}{l|cccccc|cccc}
                                  & \multicolumn{6}{c|}{\textbf{No attack}} & \multicolumn{4}{c}{\textbf{$\l_\infty$ adversarial}} \\
        \textbf{Models} & $\mathbf{mAP}$ & $\mathbf{R@1}$ & $\mathbf{imAP}$  & $\mathbf{iR@1}$  & $\mathbf{F_{mAP}}$ & $\mathbf{F_{R@1}}$  & $\mathbf{imAP}$  & $\mathbf{iR@1}$ & $\mathbf{F_{mAP}}$ & $\mathbf{F_{R@1}}$  \\
        \hline
        Undefended \cite{nguyen2021oscar} & \textbf{78.66}&\textbf{66.35} &72.83&81.05&\textbf{37.82}&\textbf{36.48} &
         15.55 &  22.87 & 12.98 & 17.01 \\
        ARIA, $\eps_\infty=\nicefrac{2}{255}$ (ours) & 54.52&39.63&78.64&85.00&32.20&27.03  &
        34.05&42.84&20.96&20.59 \\
        ARIA, $\eps_\infty=\nicefrac{4}{255}$ (ours)  & 55.86&43.38&\textbf{79.62}&85.81&32.83&28.81  &
        35.79&45.17&21.81&22.13  \\
        ARIA, $\eps_\infty=\nicefrac{8}{255}$ (ours) & 52.18&38.11&78.89&\textbf{85.88}&31.41&26.40 &
        \textbf{55.92}&\textbf{66.28}&\textbf{26.99}&\textbf{24.20}
    \end{tabular}
    \caption{Metrics for no attack and $\l_\infty$ adversarial ($\eps_\infty=\nicefrac{8}{255}$) attack for OSCAR-Net models, using queries from PSBattles. For mAP and R@1, queries have only non-editorial transforms applied. For iMAP and iR@1 digitally manipulated images with no distortions are used, both with and without adversarial perturbations. F-scores are calculated based on the appropriate mAP/R@1 following \citet{nguyen2021oscar}.}
    \label{tab:oscarnet}
        \squeezeup

\end{table*}
\subsection{Robust retrieval: \citet{nguyen2021oscar} approach}
We present our attack and defence results on OSCAR-Net \cite{nguyen2021oscar} in Table~\ref{tab:oscarnet}. Following the evaluation protocol of the attribution benchmark in \cite{nguyen2021oscar}, we report mAP/R@1 scores for the non-editorially transformed query set and imAP/iR@1 scores for the editorially transformed query set, together with the harmonic F score balancing the two terms ($F_{mAP}$ and $F_{R@1}$). Additionally, we perform adversarial attacks on the editorial query set using $\eps_\infty=\nicefrac{8}{255}$, creating a new query set called \textbf{$\l_\infty$ adversarial}. We trained  OSCAR-Net models with $\eps_\infty \in \{\nicefrac{2}{255}, \nicefrac{4}{255}, \nicefrac{8}{255}\}$ to defend against such attacks, using Eq.~(\ref{eq:objective_for_oscarnet}) and report its performance on both the no attack and \textbf{$\l_\infty$ adversarial} attack scenarios.

Whilst baseline OSCAR-Net works well, it performs poorly on \textbf{$\l_\infty$ adversarial}, with 57\% drop in imAP and iR@1 scores;  the baseline model is easily fooled by the attack. In contrast, all 3 defence models outperform OSCAR-Net by significant margins ($3\times$ improvement on imAP and iR@1 for the best model). These models also perform better at imAP and iR@1 scores on the standard benchmark, at the cost of performance reduction on the non-editorial set. We note this trade-off is already observed in the original OSCAR-Net \cite{nguyen2021oscar} and is associated with feature generalization versus discrimination. The trade-off is steered towards boosting the discrimination of editorial changes because the models are trained to defend against adversarial attacks on such changes. The value of $\eps_\infty$ can be used to determine the defence strength, with $\eps_\infty= \nicefrac{8}{255}$ yielding the best performance, closest to  performance on the standard benchmark. This is consistent with subsec.~\ref{sec:cvprw_ret_exp} when defending \cite{black2021deep}.

\subsection{Robust image comparator}
\begin{table}
    \centering
    \footnotesize
    \setlength{\tabcolsep}{3pt}
    \begin{tabular}{l|cc|cc}
    & \multicolumn{2}{c|}{\textbf{No attack}} & \multicolumn{2}{c}{\textbf{$\l_\infty$ adversarial}} \\
    \textbf{Models} & \textbf{AP} & \textbf{IoU} & \textbf{AP} & \textbf{IoU}\\    
     \hline
     Undefended ICN \citep{black2021deep} & \textbf{96.4\%} & 58.1\% & 0.6\% & 5.1\% \\  
     ARIA ICN, $\eps_\infty=\nicefrac{2}{255}$ & \textbf{96.4\%} & \textbf{61.5\%} & 65.0\% & 37.9\% \\
     ARIA ICN, $\eps_\infty=\nicefrac{4}{255}$ & 95.9\% & 59.3\% & 83.1\% & 43.7\% \\
     ARIA ICN, $\eps_\infty=\nicefrac{8}{255}$ & 95.5\% & 55.9\% & \textbf{90.7\%} & \textbf{44.9\%} \\
    \end{tabular}
    \caption{The average precision (\textbf{AP}) and intersection over union (\textbf{IoU}) between the predicted and ground truth editorial heatmaps for the \textbf{image comparator network} (ICN) with/without adversarial perturbations of radius $\eps_\infty=\nicefrac{8}{255}$.}
    \label{tab:comparator}
        \squeezeup
    \squeezeup

\end{table}
We fine-tune the model from \citet{black2021deep} for 40 epochs using 3 iterations of PGD attack for training using different $\l_\infty$ radii ($\eps_\infty \in \{\nicefrac{2}{255}, \nicefrac{4}{255}, \nicefrac{8}{255}\}$) and show the results in Table~\ref{tab:comparator}.
We benchmark robust image comparator models separately since they are trained independently of image fingerprinting models and provide complementary information about the presence of an editorial change and its location.

\paragraph{Classification module.}
First, we observe that our robust training method substantially improves the classification precision under untargeted adversarial examples. 
E.g., for the model trained with $\eps_\infty=\nicefrac{2}{255}$, we preserve the same precision as the model from \citep{black2021deep} ($96.4\%$) but substantially improve the adversarial precision (from $0.6\%$ to $65.0\%$). The most robust model is the one trained with $\eps_\infty=\nicefrac{8}{255}$ which achieves 90.7\% adversarial precision which, however, sacrifices $0.9\%$ of precision. We benchmark targeted attacks and show confusion matrix over classes in the sup. mat.

\paragraph{Heatmap module.}
Quantitatively, we improve the heatmap performance in terms of the adversarial IoU significantly: from $5.1\%$ to $44.9\%$ for the most robust model (trained with $\eps_\infty=\nicefrac{8}{255}$) which has a comparable IoU: $55.9\%$ vs. $58.1\%$. However, if we consider the robust model trained with $\eps_\infty=\nicefrac{4}{255}$, it has both better baseline IoU ($59.3\%$ vs. $58.1\%$) and adversarial IoU ($43.7\%$ vs. $5.1\%$).
Qualitative results for a robust heatmap module are visualized in Fig.~\ref{fig:attack_ic}: a robust comparator correctly highlights the same edited area regardless of whether adversarial perturbations are added. At the same time, the comparator from \citep{black2021deep} can be fooled also in a targeted way to highlight \textit{any} area (e.g., right bottom corner) as manipulated.

\section{Conclusions}
 
We began by showing that the current state-of-the-art image attribution models (both image fingerprinting and comparison), are not robust to imperceptible adversarial attacks.  This is concerning since these attacks are \textit{fully realizable} and
can be applied directly in a digital format where the attacker has \textit{white-box} access to the model.  To bridge this vulnerability, we proposed a simple and effective training technique for image attribution that significantly improves robustness to various adversarial perturbations including the ones which were unseen during training. We applied this to two fingerprinting approaches \cite{nguyen2021oscar,black2021deep}: those seeking to match manipulated content and those seeking to avoid matching such content. 
Finally, we also showed how a recent manipulation localization approach \citep{black2021deep} can be trained robustly including both its classification and heatmap modules. 

Overall, we think that adversarial vulnerability of image attribution models presents a significant negative societal impact, and that our proposed method provides a \textit{well-motivated} and \textit{practical} way to solve it. Our solution, ARIA, is particularly timely given the emergence of content authenticity standards that advocate for use of image attribution \cite{c2pa}. ARIA has two limitations: the increased training time (however, only $\approx$ $2\times$) and the trade-off for some models between a significant improvement in robustness and minor loss of accuracy (see the OSCAR-Net and image comparator experiments). Future work can further address the latter issue and additionally explore adversarial defenses for `blind' manipulation detection models \citep{Carlini_2020_CVPR_Workshops}, complementary to image attribution approaches.

\small
\bibliography{literature}
\bibliographystyle{plainnat}
\normalsize

\clearpage

\appendix
\begin{center}
	\Large\textbf{Appendix}
\end{center}

\section*{Organization of the appendix}
In Sec.~\ref{sec:training_details} we present additional training and evaluation details. 
In Sec.~\ref{sec:attack_defence_details}, we provide further implementation details for the attacks and defences both for OSCAR-Net and \citet{black2021deep} models.
In Sec.~\ref{sec:additional_exps}, we show multiple additional experiments such as accuracy of the retrieval with exact nearest neighbour search, additional hash inversion visualizations, robustness of OSCAR-Net to unseen adversarial perturbations, accuracy over classes and targeted attacks on the heatmaps for ICN models.

\section{Training and evaluation details}
\label{sec:training_details}

\paragraph{Training details.}
For the models trained according to the approach of \citet{black2021deep}, we use the learning rate $0.01$, SimCLR temperature $0.1$, 3 steps of PGD for training using step sizes $\{\nicefrac{1}{255}, \nicefrac{2}{255}, \nicefrac{4}{255}\}$ for $\eps_\infty \in \{\nicefrac{2}{255}, \nicefrac{4}{255}, \nicefrac{8}{255}\}$, respectively.

For the OSCAR-Net \citep{nguyen2021oscar} models, we use the default hyperparameters except the learning rate which is set to $1e-6$ and SimCLR temperature of $0.8$. For ARIA training, we use 3 steps of PGD with the step size $0.5 \eps_\infty$.

For the image comparator models, we use the default training hyperparameters with 3 steps of PGD for training using step sizes $\{\nicefrac{1}{255}, \nicefrac{2}{255}, \nicefrac{4}{255}\}$ for $\eps_\infty \in \{\nicefrac{2}{255}, \nicefrac{4}{255}, \nicefrac{8}{255}\}$, respectively.

\paragraph{Evaluation details.}
For the attacks unseen during training, we use $200$ iterations of PGD (we increase it from $50$ iterations used throughout the paper to account for larger perturbation radii) using the step size of $\eps_\infty=\nicefrac{4}{255}$ for $\ell_\infty$-perturbations and $\eps_2=0.5$ for $\ell_2$-perturbations.

For hash inversions, we use 1000 iterations of PGD with the step size $\nicefrac{4}{255}$, and the approximation parameter $\beta=1$.

\paragraph{Training time.}
Standard training of the \citet{black2021deep} model on Behance1M takes 34.3 hours while ARIA training takes 72.8 hours (i.e., $2.3\times$ factor slowdown) on two NVIDIA V100 GPUs for 20 epochs. 

Standard OSCAR-Net training on PSBattles takes 31.6 hours while ARIA training takes 65.1 hours (i.e., $2.1\times$ factor slowdown) on a single NVIDIA GeForce RTX 3090 GPU for 10 epochs. %

We note that for both models, ARIA uses 3 steps of PGD for training but the slowdown factor is less than $4\times$ which is due to more effective GPU utilization for robust training.

\begin{figure*}[t]
    \centering
    \begin{tabular}{c}
        \textbf{OSCAR-Net} \cite{nguyen2021oscar} \\
        \includegraphics[width=1.0\linewidth]{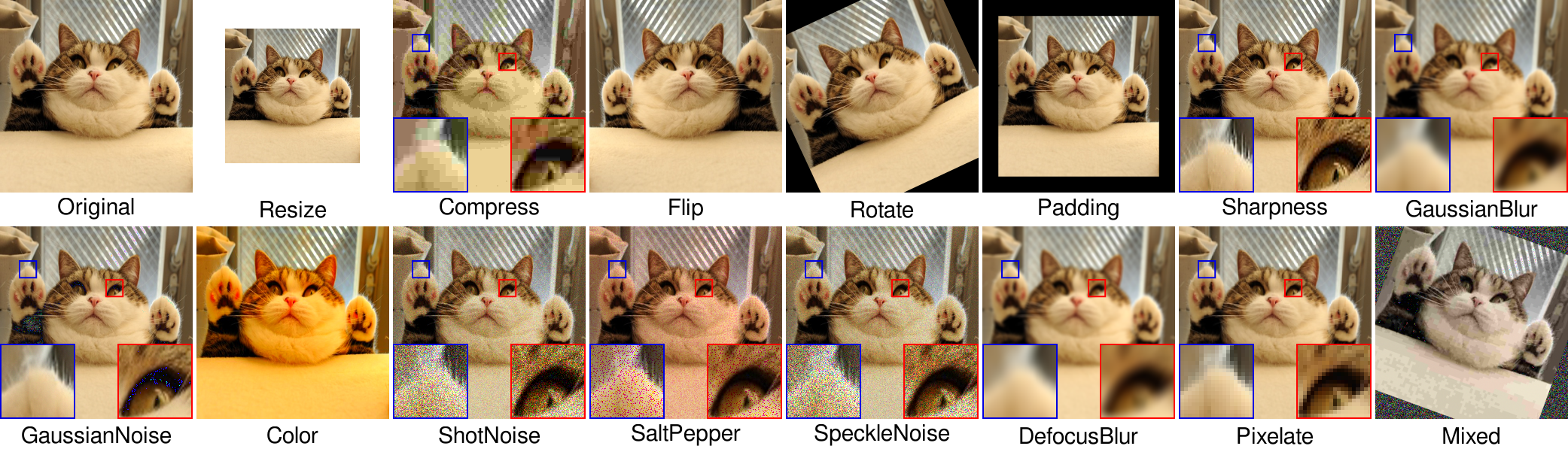} \\
        \textbf{\citet{black2021deep}} \\
        \includegraphics[width=1.0\linewidth]{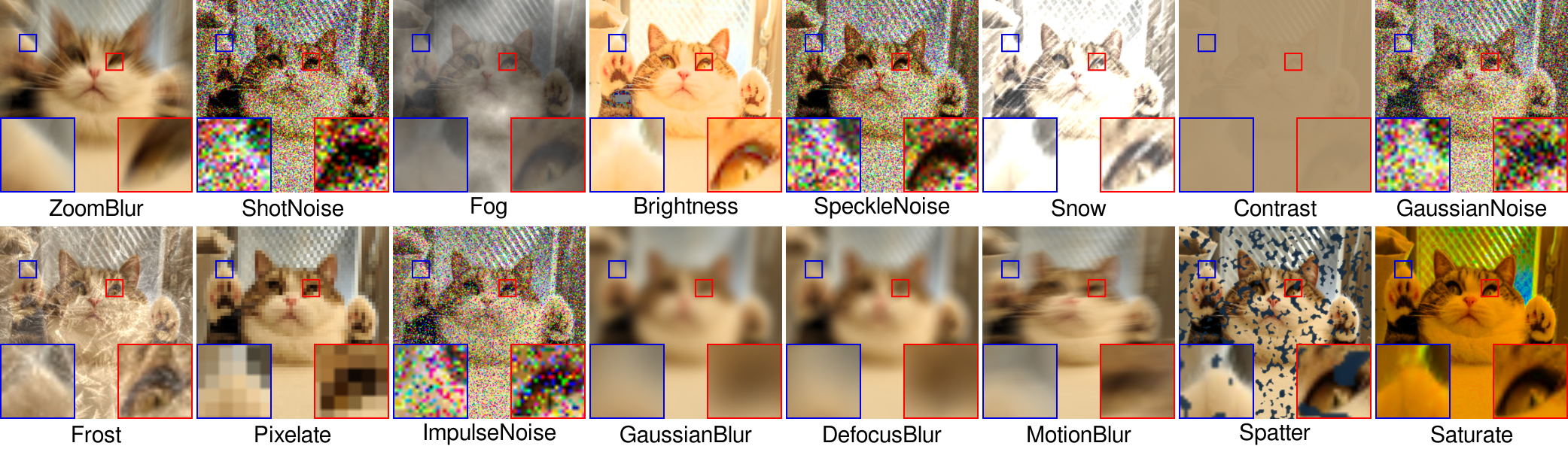}
    \end{tabular}
    \caption{Examples of non-editorial changes applied to the same image from PSBattles according to the query sets used to evaluate the OSCAR-Net \cite{nguyen2021oscar} and Black \etal \cite{black2021deep} approaches.}
    \label{fig:noneditorial_changes_single_image}
\end{figure*}

\begin{figure*}[t!]
    \centering
    \includegraphics[width=0.16\textwidth]{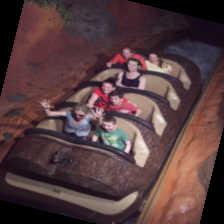}
    \includegraphics[width=0.16\textwidth]{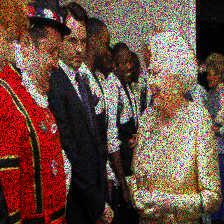}
    \includegraphics[width=0.16\textwidth]{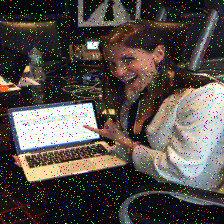}
    \includegraphics[width=0.16\textwidth]{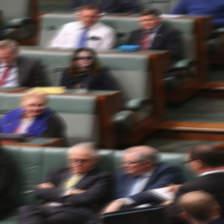}
    \includegraphics[width=0.16\textwidth]{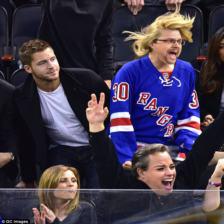}
    \includegraphics[width=0.16\textwidth]{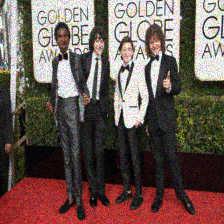}
    \\
    \includegraphics[width=0.16\textwidth]{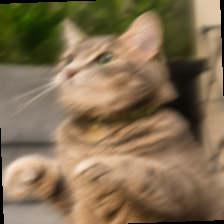}
    \includegraphics[width=0.16\textwidth]{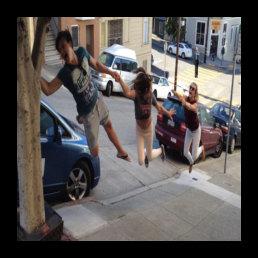}
    \includegraphics[width=0.16\textwidth]{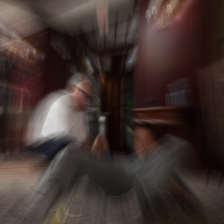}
    \includegraphics[width=0.16\textwidth]{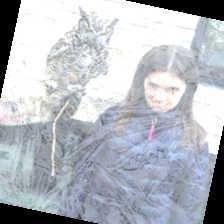}
    \includegraphics[width=0.16\textwidth]{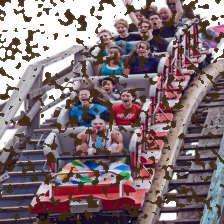}
    \includegraphics[width=0.16\textwidth]{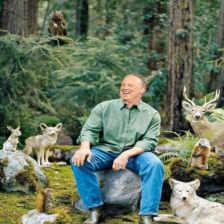}
    \\
    \includegraphics[width=0.16\textwidth]{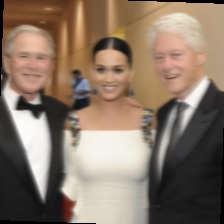}
    \includegraphics[width=0.16\textwidth]{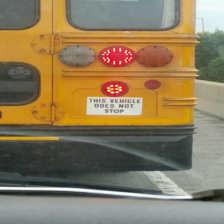}
    \includegraphics[width=0.16\textwidth]{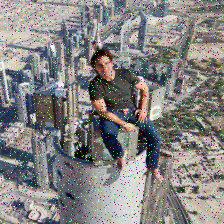}
    \includegraphics[width=0.16\textwidth]{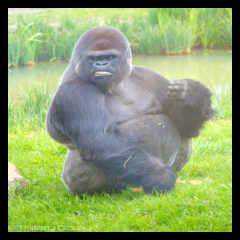}
    \includegraphics[width=0.16\textwidth]{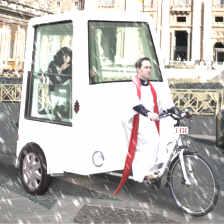}
    \includegraphics[width=0.16\textwidth]{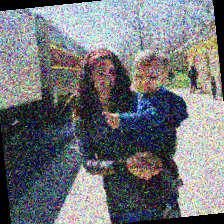}
    \caption{Additional examples of non-editorial changes applied to the images from PSBattles.}
    \label{fig:noneditorial_changes_examples}
    \vspace{1em}
\end{figure*}

\paragraph{Examples of non-editorial transformations.}
In Fig.~\ref{fig:noneditorial_changes_single_image} and Fig.~\ref{fig:noneditorial_changes_examples}, we show images with non-editorial changes from PSBattles which we used for the  ``\textit{Editorial + non-editorial}'' query sets for evaluation of the OSCAR-Net models and models of \citet{black2021deep}. %

\begin{table*}[t]
    \centering
    \footnotesize
    \setlength\tabcolsep{2.5pt}
    \begin{tabular}{l|cccc|cccc|cccc}

         \multicolumn{1}{c}{} & \multicolumn{12}{c}{\textbf{Top-1 and top-100 recall for different query sets}} \\
         & \multicolumn{4}{c|}{\textbf{Non-editorial distortions}} & \multicolumn{4}{c|}{\textbf{Editorial manipulations}} & \multicolumn{4}{c}{\textbf{Editorial + non-editorial}} \\
                              & \multicolumn{2}{c}{\textbf{No attack}} & \multicolumn{2}{c|}{\textbf{$\l_\infty$ adversarial}}  &  \multicolumn{2}{c}{\textbf{No attack}} & \multicolumn{2}{c|}{\textbf{$\l_\infty$ adversarial}}  &  \multicolumn{2}{c}{\textbf{No attack}} & \multicolumn{2}{c}{\textbf{$\l_\infty$ adversarial}} \\
        \textbf{Existing models}  & \textbf{R@1} & \textbf{R@100} & \textbf{R@1} & \textbf{R@100}  &  \textbf{R@1} & \textbf{R@100} & \textbf{R@1} & \textbf{R@100}  &  \textbf{R@1} & \textbf{R@100} & \textbf{R@1} & \textbf{R@100}  \\
        \hline
        Standard supervised, ImageNet \citep{paszke2017automatic}  &  45.1 & 59.3 & 0.0 & 0.2  &  98.3 & \textbf{99.6} & 0.1 & 0.3  &  37.3 & 52.9 & 0.0 & 0.3 \\
        DeepAugment + AugMix supervised, ImageNet \citep{Hendrycks2020Many}  &  75.2 & 84.5 & 0.2 & 2.0  &  \textbf{98.5} & \textbf{99.6} & 0.0 & 0.6  &  67.8 & 80.7 & 0.0 & 0.3 \\
        Robust supervised, $\eps_\infty=\nicefrac{4}{255}$, ImageNet \citep{Salman2020Do}  &  57.3 & 66.1 & 30.3 & 44.0  &  97.4 & 99.2 & 79.7 & 92.4  &  51.2 & 62.0 & 22.4 & 38.0 \\
        Undefended contrastive, PSBattles \citep{black2021deep}  &  86.2 & 96.7 & 0.0 & 0.0  &  87.7 & 95.5 & 0.0 & 0.0  &  70.0 & 89.5 & 0.0 & 0.0  \\
        
        \textbf{Our new models} \\
        \hline
        Undefended contrastive, Behance &  99.2 & 99.9 & 4.8 & 25.3  &  94.4 & 97.6 & 0.9 & 9.8  &  91.9 & 96.8 & 2.6 & 16.1 \\
        ARIA contrastive + hashing, $\eps_\infty=\nicefrac{4}{255}$, Behance &  96.8 & 98.7 & 83.8 & 89.3  &  92.1 & 96.7 & 85.2 & 93.8  &  87.1 & 94.5 & 69.2 & 82.7 \\
        ARIA contrastive + hashing, $\eps_\infty=\nicefrac{8}{255}$, Behance &  93.5 & 96.5 & 84.1 & 90.8  &  91.4 & 96.0 & 87.0 & 93.9  &  82.8 & 91.1 & 69.7 & 82.4 \\
        ARIA contrastive, $\eps_\infty=\nicefrac{2}{255}$, Behance &  \textbf{99.5} & \textbf{100.0} & 87.7 & 90.7  &  96.1 & 98.6 & 91.6 & 96.9  &  \textbf{94.8} & \textbf{98.1} & 78.6 & 87.3 \\
        ARIA contrastive, $\eps_\infty=\nicefrac{4}{255}$, Behance &  99.4 & 99.9 & 90.5 & 92.7  &  96.1 & 98.4 & \textbf{93.4} & \textbf{97.3}  &  94.7 & 97.9 & \textbf{83.3} & 90.4 \\
        ARIA contrastive, $\eps_\infty=\nicefrac{8}{255}$, Behance &  98.6 & 99.7 & \textbf{94.5} & \textbf{95.4}  &  95.5 & 98.3 & 93.2 & 97.1  &  92.8 & 97.2 & 82.9 & \textbf{90.9} \\
    \end{tabular}
    \caption{Standard and $\l_\infty$ adversarial ($\eps_\infty=\nicefrac{8}{255}$) top-1 and top-100 recall for different ResNet-50 models evaluated on PSBattles \citep{heller2018ps}. The database contains original images from PSBattles and 2M distractor images from Stock indexed using the \textbf{exact nearest neighbour search} (unlike Table 1 in the main part that used the approximate IVF1024, PQ16 index). We use three query sets based on PSBattles: (1) non-editorial distortions (ImageNet-C and affine) on original images, (2)  editorial manipulations but no distortions, (3)  editorial manipulations with non-editorial distortions.}
    \label{tab:recall_cvprw_models_exact_index}
\end{table*}

\section{Further details on the attack and defence scope on OSCAR-Net and Black \etal models}
\label{sec:attack_defence_details}
A model needs to be differentiable with respect to the input image in order to perform an effective adversarial attack (and defence) on it. In other words, our main prerequisite is that we should be able to back-propagate the gradient of the loss to the original input. Despite being complex attribution models, we show that OSCAR-Net \cite{nguyen2021oscar} and Black \etal \cite{black2021deep} both can meet this requirement.

\noindent \textbf{OSCAR-Net} consists of an object detection module (Mask-RCNN \cite{maskrcnn}) to decompose an image into a set of objects, followed by 3 sub-networks to learn the global image features, object-level features (including object CNN visual, shape and geometry features) as well as the relation features between objects. These features are pooled via a fully-connected graph transformer network to produce a compact binary embedding. Note that OSCAR-Net does not aim to learn object detection (the Mask-RCNN module weights are not updated during training), and we do the same. Here we focus on attacking and defending the multi-branch feature extraction and aggregation which are learnable in OSCAR-Net. Thus, we apply our perturbations to the full image after the object detection step, i.e. we treat the output of the object detector as constant. We note that there exists adversarial attack and defence approaches on object detection \cite{chen2021class} and integrating those on OSCAR-Net could be a topic of future work.

\noindent \textbf{Black \etal} consists of two distinct models that are trained separately: an image retrieval model insensitive to both editorial and non-editorial changes, followed by an image comparator (IC) model distinguishing editorial from non-editorial transformations. Given a query, the image retrieval model returns top-k candidate images which are brought to the IC model to determine if there exists a `matched' image among the candidates and whether the query has editorial or non-editorial changes. The IC model also outputs an editorial heatmap if editorial change is predicted on a query-candidate pair. The retrieval model has a simple ResNet-50 architecture and is trained with SimCLR loss \cite{black2021deep}, hence is fully differentiable. The IC model is more complex with a dewarping unit to align the query with the candidate image, followed by a CNN-based feature extraction module to output the editorial prediction and heatmap. Both sub-modules are differentiable with respect to the input image pair and we have demonstrated that adversarial attacks could be performed on both prediction and heatmap in our main paper, as well as an adversarially robust training method to defend against such attacks.

We refer to \cite{black2021deep,nguyen2021oscar} for more details on the architecture and training strategies of the two above approaches.

\section{Additional experiments}
\label{sec:additional_exps}

\begin{table*}[t!]
    \centering
    \footnotesize
    \setlength\tabcolsep{2.5pt}
    
    \begin{tabular}{l|cccc|cccc|cccc}
                                  & \multicolumn{4}{c}{\textbf{$\l_\infty$ adversarial, $\epsilon_\infty=\nicefrac{16}{255}$}} & \multicolumn{4}{|c|}{\textbf{$\l_\infty$ adversarial, $\epsilon_\infty=\nicefrac{32}{255}$}} & \multicolumn{4}{c}{\textbf{$\l_2$ adversarial, $\epsilon_2=5$}} \\
        \textbf{Models} & $\mathbf{imAP}$  & $\mathbf{iR@1}$ & $\mathbf{F_{mAP}}$ & $\mathbf{F_{R@1}}$ & $\mathbf{imAP}$  & $\mathbf{iR@1}$ & $\mathbf{F_{mAP}}$ & $\mathbf{F_{R@1}}$ & $\mathbf{imAP}$  & $\mathbf{iR@1}$ & $\mathbf{F_{mAP}}$ & $\mathbf{F_{R@1}}$  \\
        \hline
        Undefended \cite{nguyen2021oscar}	&7.69	&11.08	&7.01	&9.50	&5.84	&8.18	&5.44	&7.28	&38.04	&45.37	&25.64	&\textbf{26.95}\\
ARIA, $\eps_\infty=\nicefrac{2}{255}$ (ours)	&22.64	&29.93	&16.00	&17.05	&17.09	&23.07	&13.01	&14.58	&\textbf{54.30}	&\textbf{61.55}	&\textbf{27.21}	&24.11\\
ARIA, $\eps_\infty=\nicefrac{4}{255}$ (ours)	&21.04	&27.97	&15.29	&17.01	&16.76	&22.36	&12.89	&14.76	&47.46	&55.44	&25.66	&24.34\\
ARIA, $\eps_\infty=\nicefrac{8}{255}$ (ours)	&\textbf{41.85}	&\textbf{49.56}	&\textbf{23.22}	&\textbf{21.54}	&\textbf{40.43}	&\textbf{47.09}	&\textbf{22.78}	&\textbf{21.06}	&42.14	&51.52	&23.31	&21.91\\
    \end{tabular}
    \caption{Performance metrics for attacks \textit{unseen} during training for OSCAR-Net models, using queries from PSBattles. Evaluation is on a query set of digitally manipulated images with no distortions.}
    \label{tab:oscarnet_unseen}
\end{table*}
\begin{table*}[t!]
    \centering
    \footnotesize
    \setlength{\tabcolsep}{3pt}
    \begin{tabular}{l|cccc|cccc}
    & \multicolumn{4}{c|}{\textbf{Average precision, no attack}} & \multicolumn{4}{c}{\textbf{Average precision, $\l_\infty$ adversarial attack}} \\
    \textbf{Models} & \textbf{All} & \textbf{Non-editorial} & \textbf{Edit. + non-} & \textbf{Different} & \textbf{All} & \textbf{Non-editorial} & \textbf{Edit. + non-} & \textbf{Different} \\    
                    & \textbf{classes} & \textbf{changes} & \textbf{edit. changes} & \textbf{images} & \textbf{changes} & \textbf{changes} & \textbf{edit. changes} & \textbf{images} \\    
     \hline
     Undefended ICN \citep{black2021deep} & \textbf{96.4\%} & \textbf{98.2\%} & 91.4\% & 99.6\%  &  0.6\% & 0.0\% & 0.1\% & 1.6\% \\  
     ARIA ICN, $\eps_\infty=\nicefrac{2}{255}$ & \textbf{96.4\%} & 91.8\% & \textbf{97.7\%} & \textbf{99.7\%}  &  65.0\% & 21.6\% & 84.9\% & 85.6\% \\
     ARIA ICN, $\eps_\infty=\nicefrac{4}{255}$ & 95.9\% & 91.6\% & 97.0\% & 99.3\%  &  83.1\% & 67.6\% & 87.1\% & 93.9\% \\
     ARIA ICN, $\eps_\infty=\nicefrac{8}{255}$ & 95.5\% & 92.2\% & 95.5\% & 98.5\%  &  \textbf{90.7\%} & \textbf{86.6\%} & \textbf{88.7\%} & \textbf{96.2\%} \\
    \end{tabular}
    \caption{The average precision for the \textbf{image comparator network} (ICN) with/without adversarial perturbations of radius $\eps_\infty=\nicefrac{8}{255}$ over three different classes (depending on the query image that can be either the same image with non-editorial changes, the same image with editorial and non-editorial changes, or a different image).}
    \label{tab:comparator_precision_extra}
\end{table*}

\paragraph{Retrieval with exact nearest neighbour search for \citet{black2021deep} models.}
First of all, we note that exact nearest neighbour search reported in Table~\ref{tab:recall_cvprw_models_exact_index} is not practical for databases that contain millions of images and we report it so that we can analyze the performance drop which occurs due to approximate image retrieval. 
Table~\ref{tab:recall_cvprw_models_exact_index} suggests that overall the trends and rankings between different methods are the same as in Table~1 from the main part of the paper. At the same time, as expected, the absolute numbers are higher: e.g., standard top-1 recall for the ARIA model trained with $\eps_\infty=\nicefrac{8}{255}$ is 99.5\% compared to 97.3\% with the approximate indexing reported in the main part. Such performance drop is uniform over different methods.
We can also see that ImageNet-trained models perform well on images with editorial changes. However, we note that the ImageNet models use the embedding dimension of $2048$ which is much larger the $256$ used by our contrastively trained models and leads to even slower search time.

\paragraph{Robustness of OSCAR-Net models to unseen adversarial perturbations.}
Table~\ref{tab:oscarnet_unseen} shows the robustness results of OSCAR-Net for perturbations which were unseen during training. These are $\l_2$-bounded perturbations ($\eps_2 = 5$) and $\l_\infty$-perturbations of a larger radius compared to those used for training ($\eps_\infty \in \{\nicefrac{16}{255}, \nicefrac{32}{255}\}$).

The robustness generalises very well to the larger $\l_\infty$-perturbations: e.g. with perturbations of size $\eps_\infty = \nicefrac{32}{255}$ the $F_{mAP}$ score for the undefended model of \citet{nguyen2021oscar} is reduced to 5.44\%, but for all our defended models it is at least 12.89\%. In the case of our best defended model it is 22.78\%. The $\l_2$ perturbations with $\eps_2 = 5$ are not very successful at attacking the OSCAR-Net model, so it is not possible to draw conclusions about robustness in this case. We think that for $\l_2$ perturbations treating the object detector's output as constant can be suboptimal but we leave better attacks tailored to the OSCAR-Net architecture to future work.

\paragraph{Image comparator models: accuracy over classes.}
We show the results in Table~\ref{tab:comparator_precision_extra} where we report the average precision over three classes depending on the query image that can be either the same image with non-editorial changes, the same image with editorial and non-editorial changes, or a different image. 
We can see that the standard precision is approximately uniform over different classes but the adversarial precision can be non-uniform. For example, the ARIA ICN model trained with $\eps_\infty=\nicefrac{2}{255}$ has only 21.6\% adversarial precision on the same images with non-editorial changes. However, using a higher $\eps$ for ARIA fixes this problem, e.g., for $\eps_\infty=\nicefrac{8}{255}$ we get 86.6\% adversarial precision.

\paragraph{Image comparator models: targeted attacks on heatmaps.}
We show the results of targeted attacks on the image comparator models in Table~\ref{tab:comparator_iou_targeted}.
For the attack, we target a random cell of a $7\times7$ heatmap by maximizing the cosine loss.
We note that unlike other metrics, a lower targeted intersection over union (IoU) is better as it implies a smaller overlap of the predicted heatmap with the wrong target heatmap. 
We can observe that ARIA training successfully reduces the success rate of the attack in terms of IoU from 48.3\% (undefended ICN) down to 3.9\% (ARIA training with $\eps_\infty=\nicefrac{8}{255}$).

\begin{table}[t!]
    \centering
    \footnotesize
    \setlength{\tabcolsep}{3pt}
    \begin{tabular}{l|c|c}
    & \multicolumn{1}{c|}{\textbf{No attack}} & \multicolumn{1}{c}{\textbf{$\l_\infty$ adversarial}} \\
    \textbf{Models} & \textbf{IoU} & \textbf{Targeted IoU} \\    
     \hline
     Undefended ICN \citep{black2021deep} & 58.1\% & 48.3\% \\  
     ARIA ICN, $\eps_\infty=\nicefrac{2}{255}$ & \textbf{61.5\%} & 10.0\% \\
     ARIA ICN, $\eps_\infty=\nicefrac{4}{255}$ & 59.3\% & 5.4\% \\
     ARIA ICN, $\eps_\infty=\nicefrac{8}{255}$ & 55.9\% & \textbf{3.9\%} \\
    \end{tabular}
    \caption{The average intersection over union (\textbf{IoU}) between the predicted and ground truth editorial heatmaps for the \textbf{image comparator network} (ICN) with/without \textit{targeted} adversarial perturbations of radius $\eps_\infty=\nicefrac{8}{255}$. Note that unlike other metrics, a lower targeted IoU is better as it implies a smaller overlap of the predicted heatmap with the wrong target heatmap.}
    \label{tab:comparator_iou_targeted}
\end{table}

\begin{figure*}[t!]
    \centering
    \scriptsize
    \begin{subfigure}[c]{.15\textwidth}
        \caption{\scriptsize \textbf{Original image $x$}}
        \includegraphics[width=\textwidth]{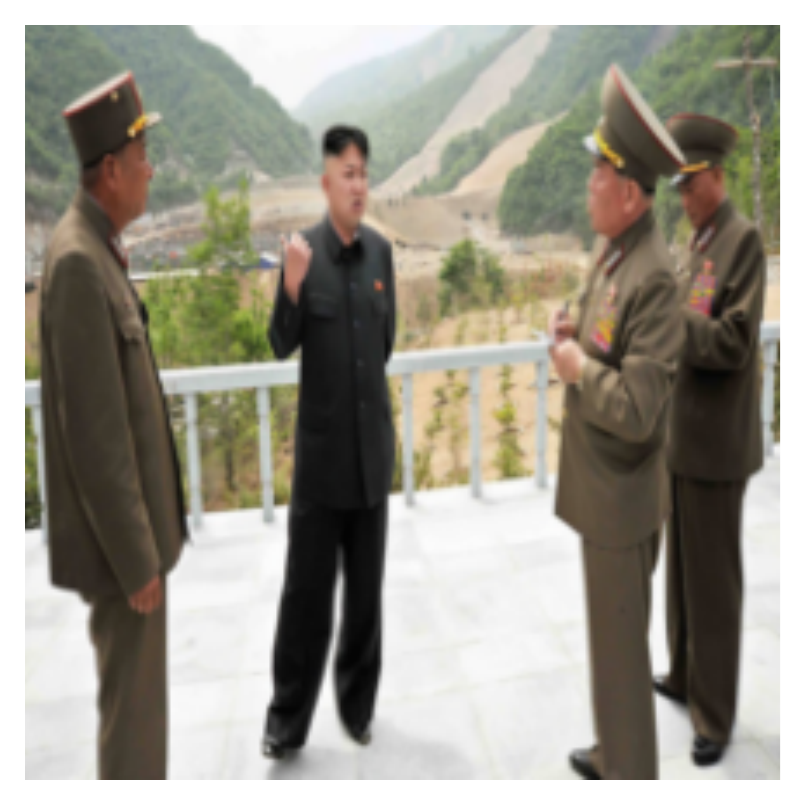}
        \includegraphics[width=\textwidth]{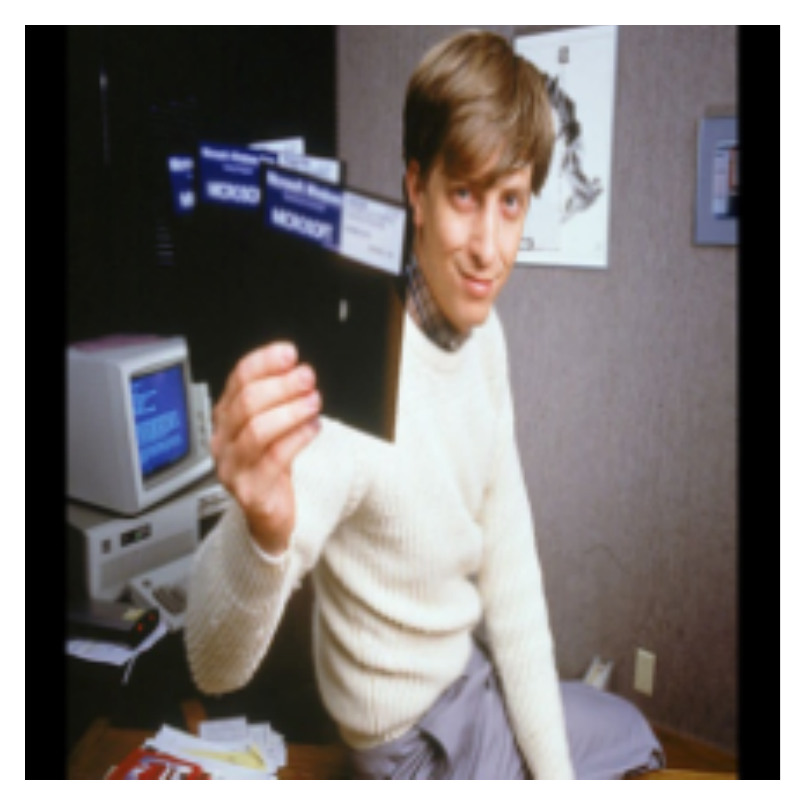}
    \end{subfigure}
    \begin{subfigure}[c]{.15\textwidth}
        \caption{\scriptsize \textbf{$\phi^{-1}_{\approx}(x)$, standard model}}
        \includegraphics[width=\textwidth]{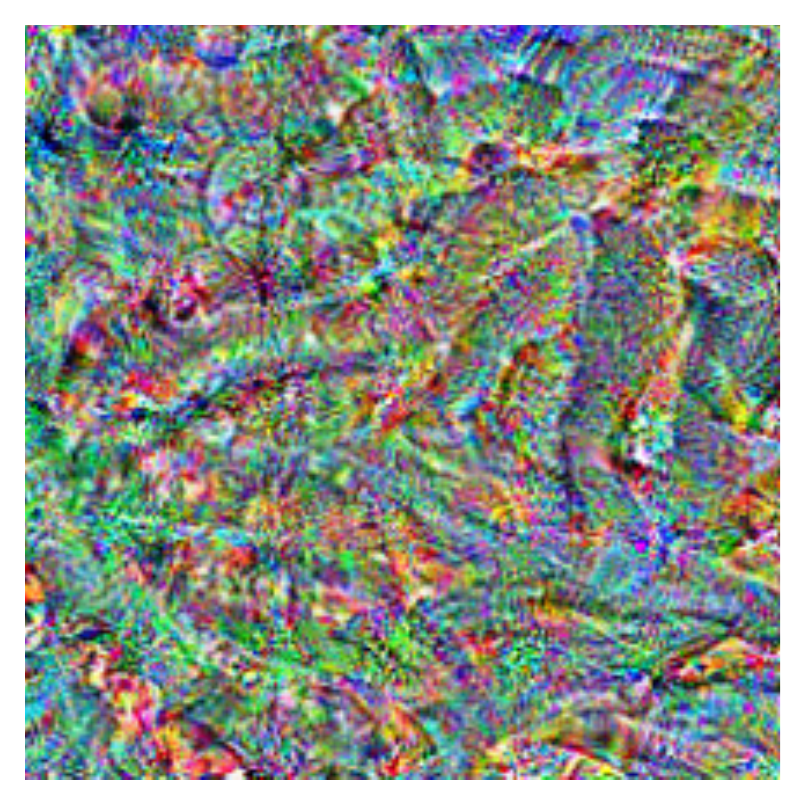}
        \includegraphics[width=\textwidth]{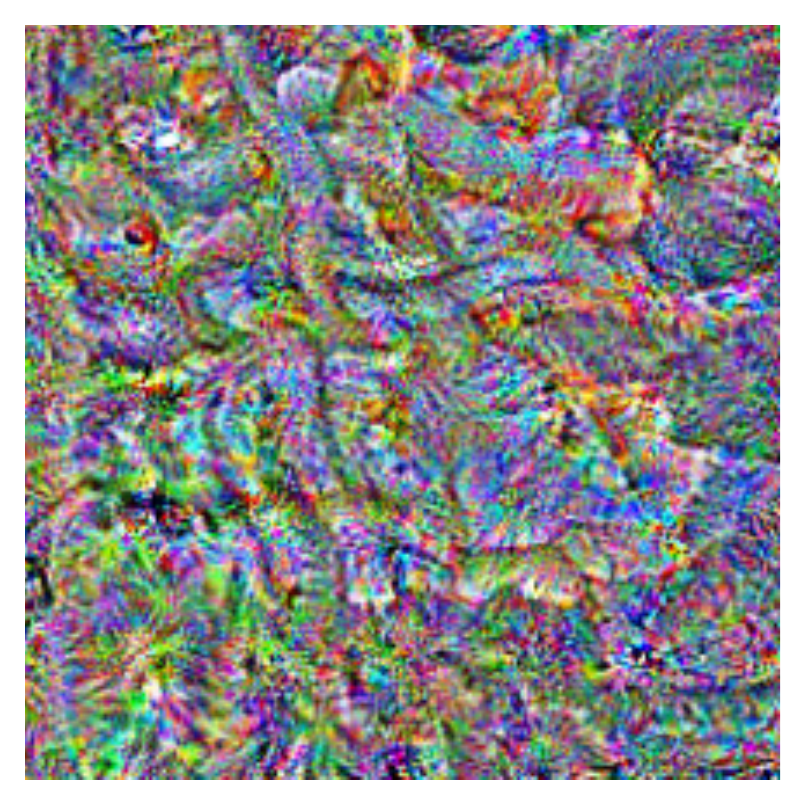}
    \end{subfigure}
    \begin{subfigure}[c]{.15\textwidth}
        \caption{\scriptsize \textbf{$\phi^{-1}_{\approx}(x)$, robust model}}
        \includegraphics[width=\textwidth]{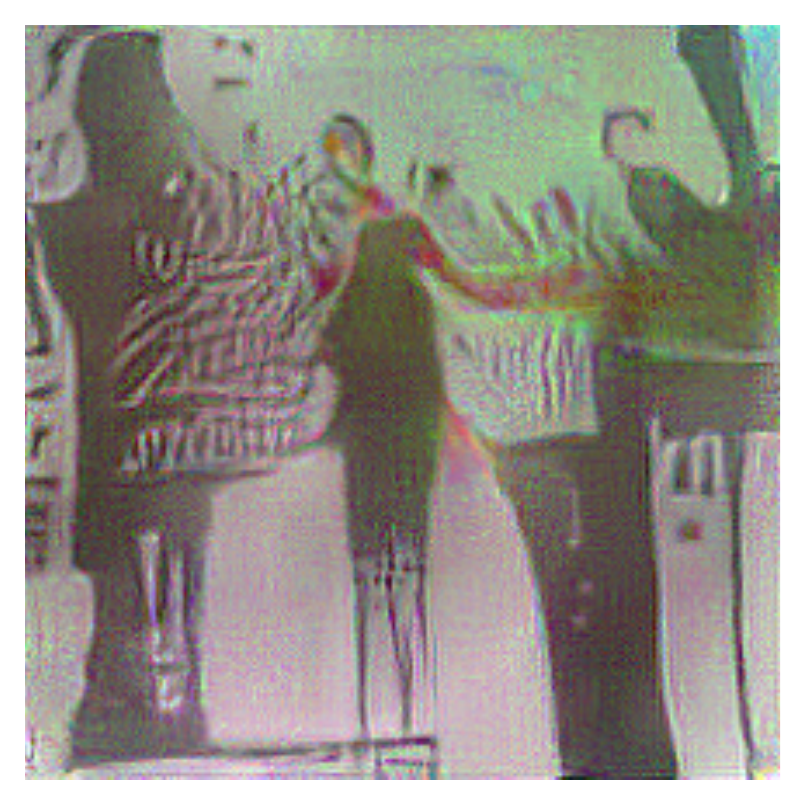}
        \includegraphics[width=\textwidth]{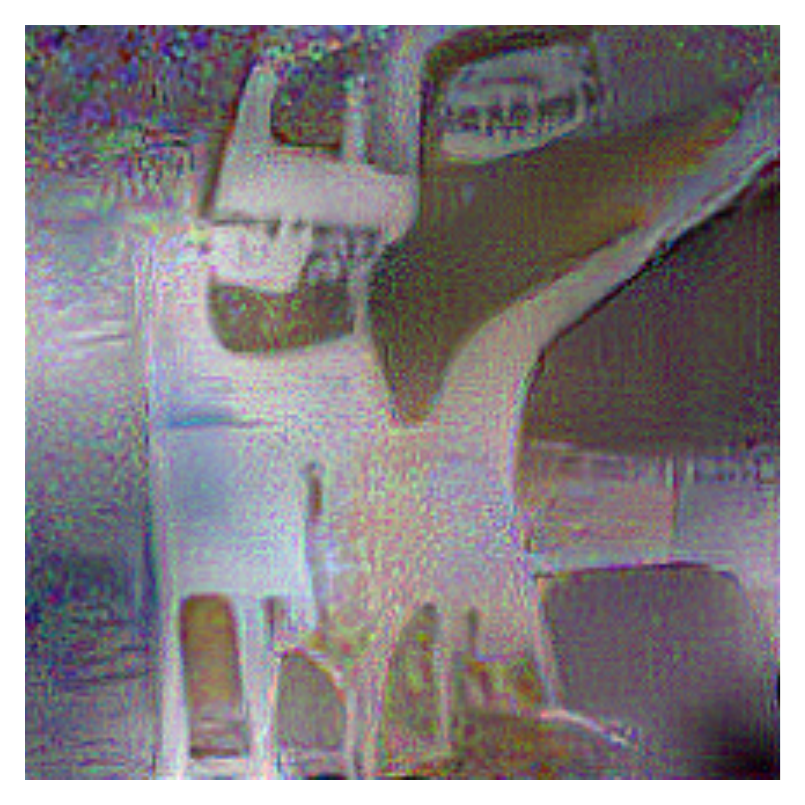}
    \end{subfigure}
    \hspace{5mm}
    \begin{subfigure}[c]{.15\textwidth}
        \caption{\scriptsize \textbf{Original image $x$}}
        \includegraphics[width=\textwidth]{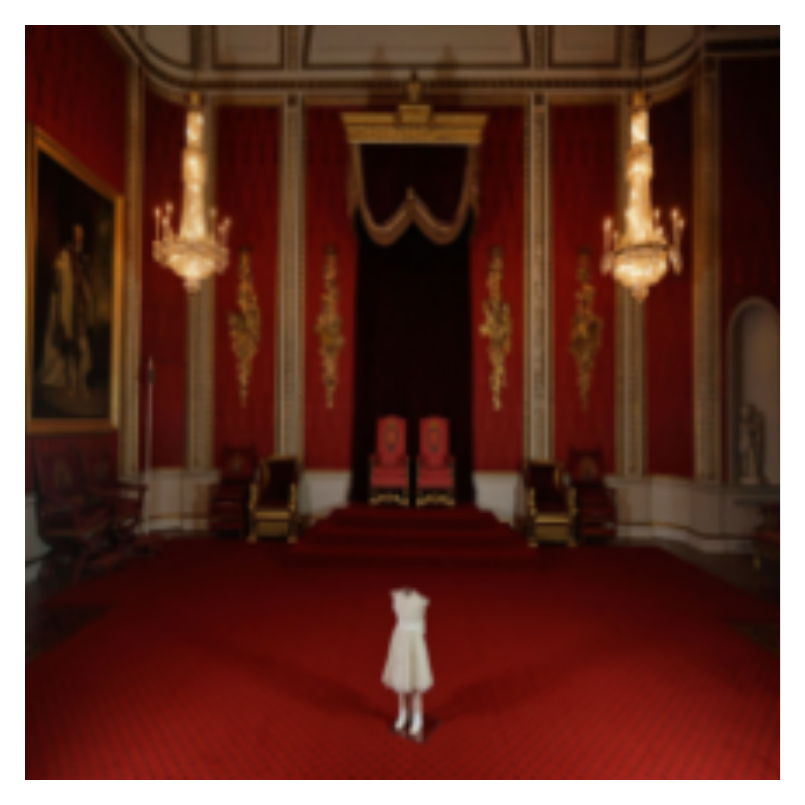}
        \includegraphics[width=\textwidth]{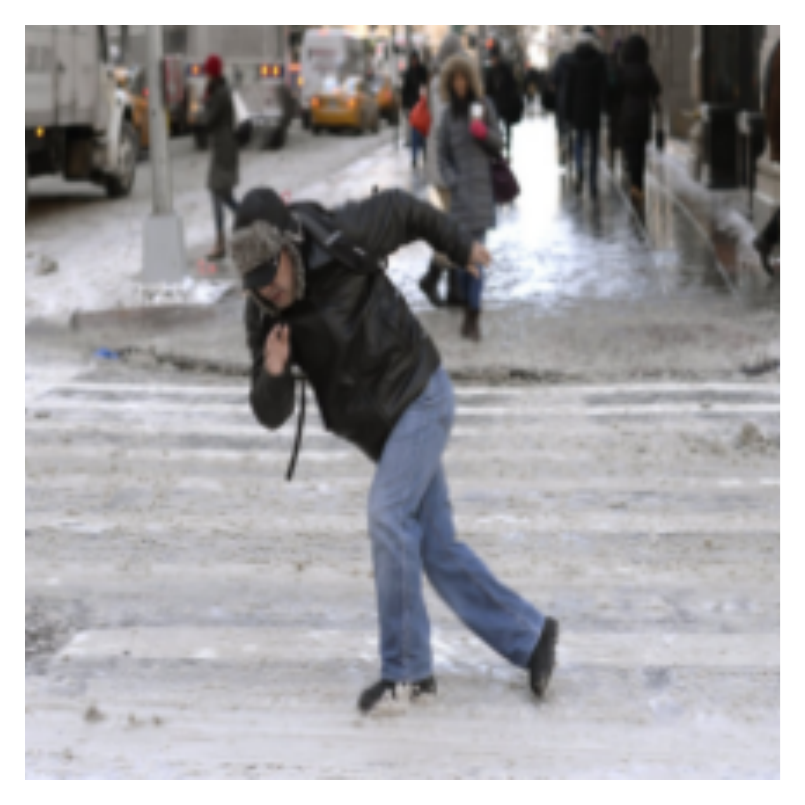}
    \end{subfigure}
    \begin{subfigure}[c]{.15\textwidth}
        \caption{\scriptsize \textbf{$\phi^{-1}_{\approx}(x)$, standard model}}
        \includegraphics[width=\textwidth]{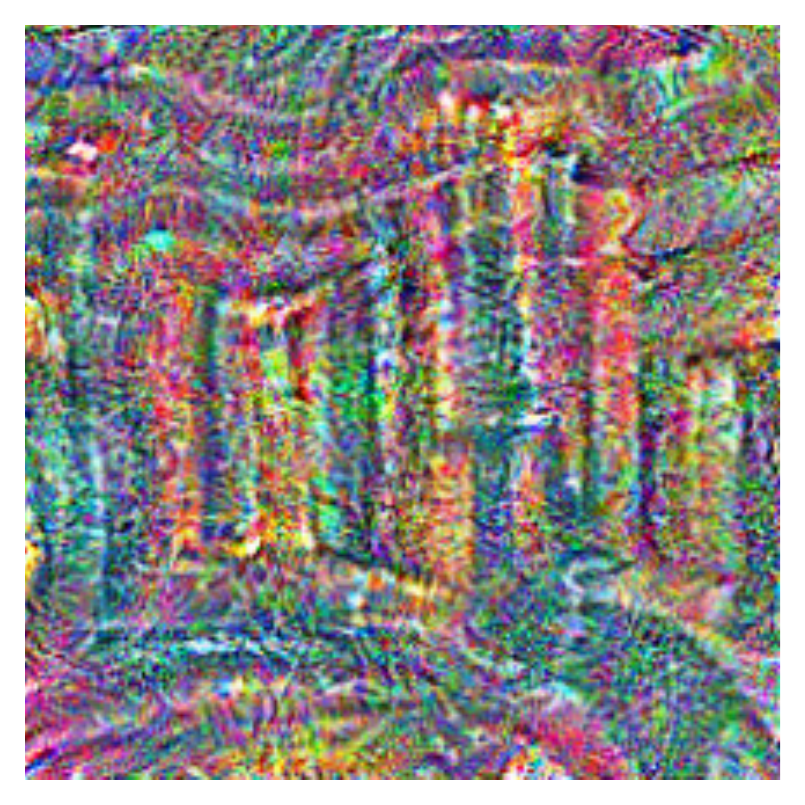}
        \includegraphics[width=\textwidth]{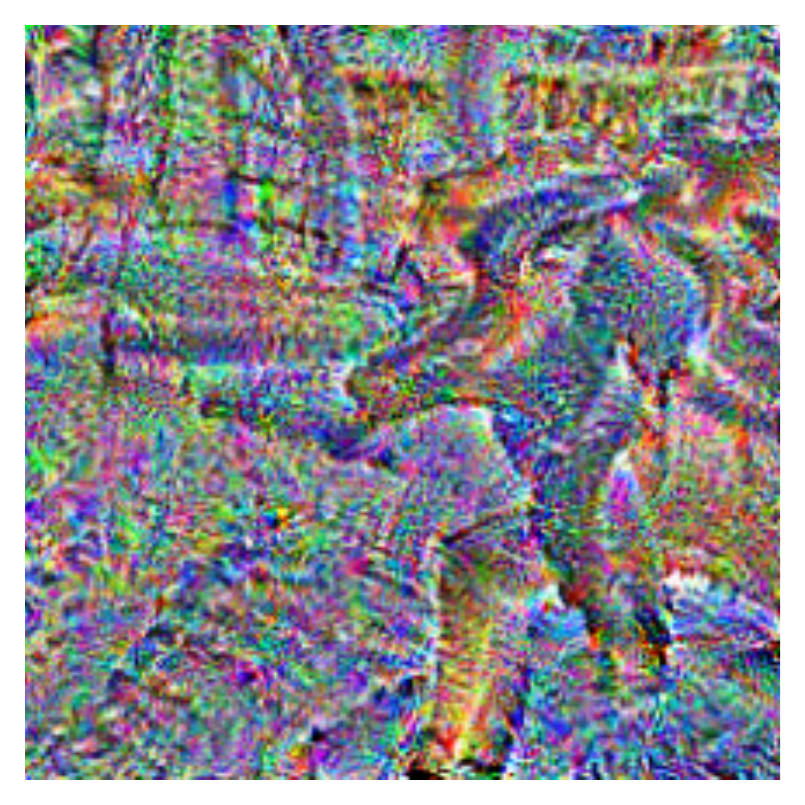}
    \end{subfigure}
    \begin{subfigure}[c]{.15\textwidth}
        \caption{\scriptsize \textbf{$\phi^{-1}_{\approx}(x)$, robust model}}
        \includegraphics[width=\textwidth]{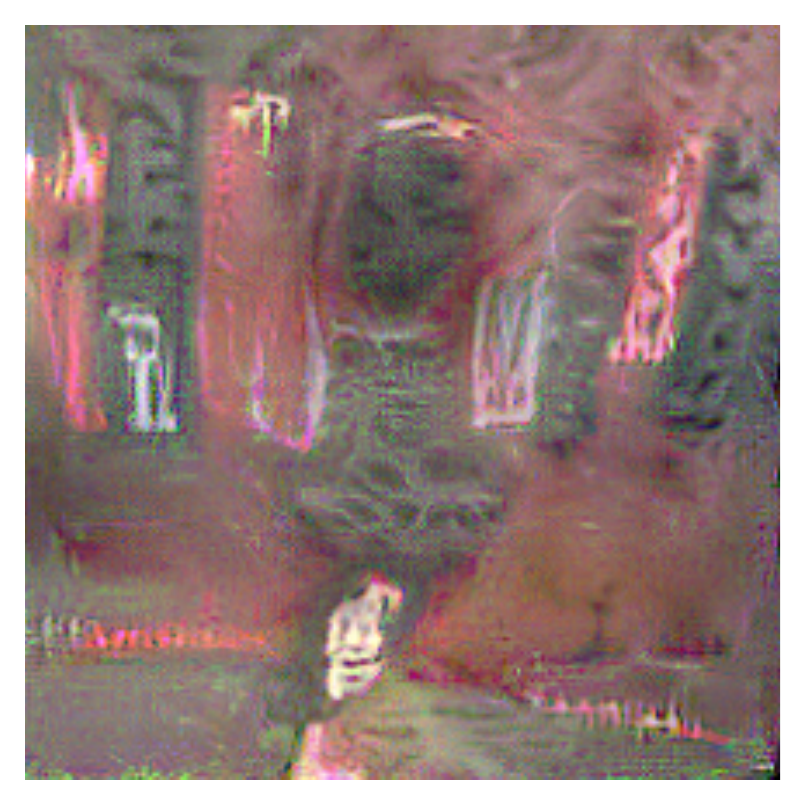}
        \includegraphics[width=\textwidth]{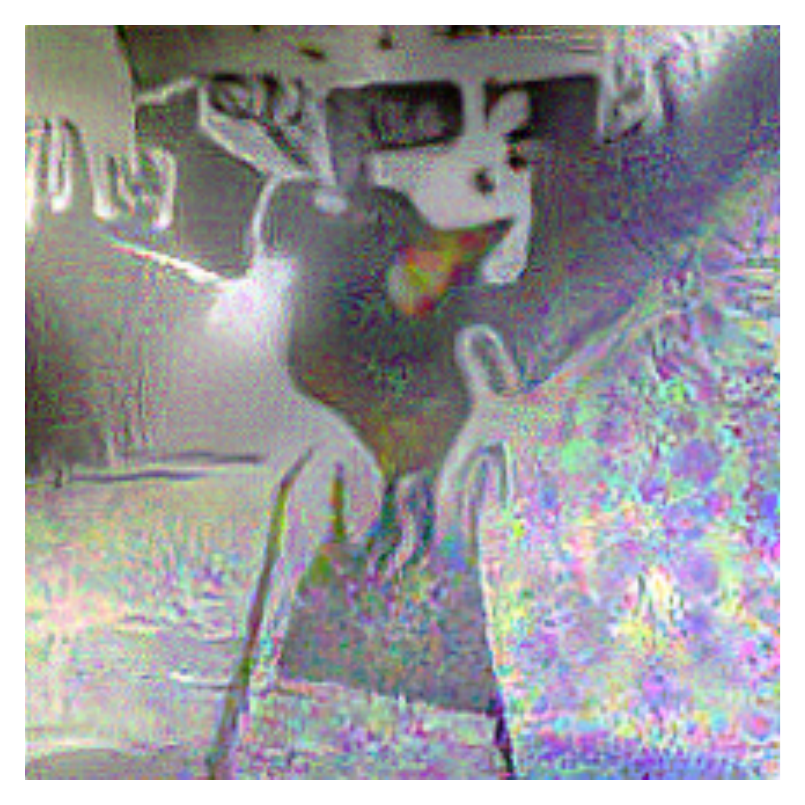}
    \end{subfigure}
    \\
    \begin{subfigure}[c]{.15\textwidth}
        \includegraphics[width=\textwidth]{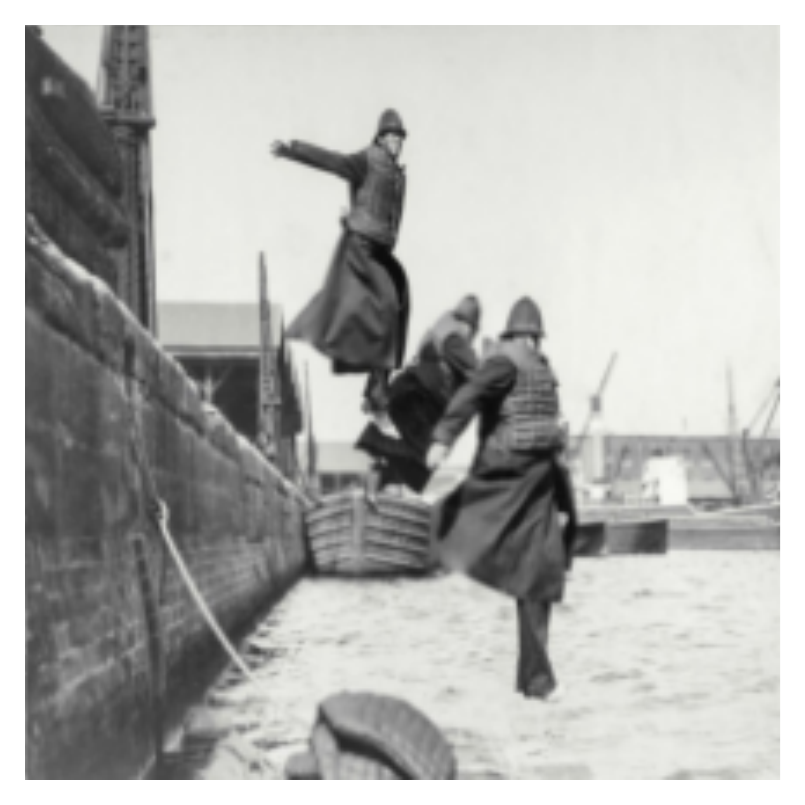}
        \includegraphics[width=\textwidth]{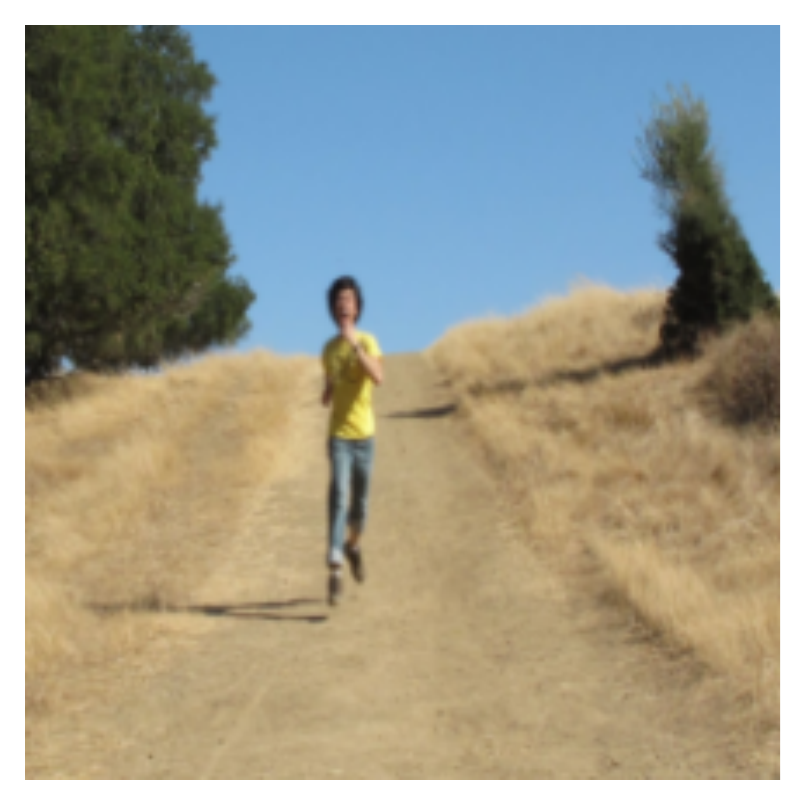}
    \end{subfigure}
    \begin{subfigure}[c]{.15\textwidth}
        \includegraphics[width=\textwidth]{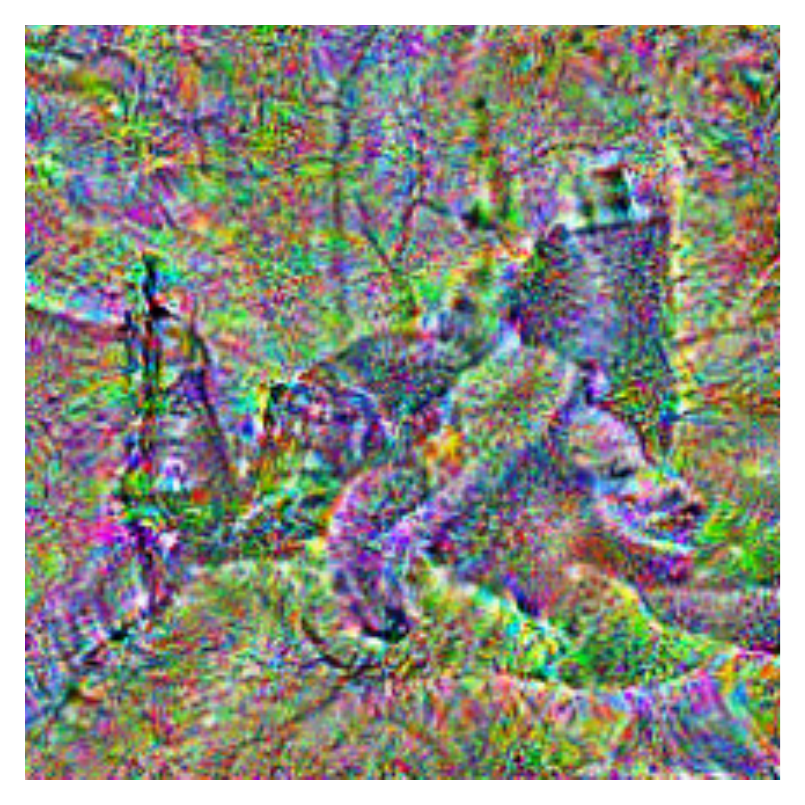}
        \includegraphics[width=\textwidth]{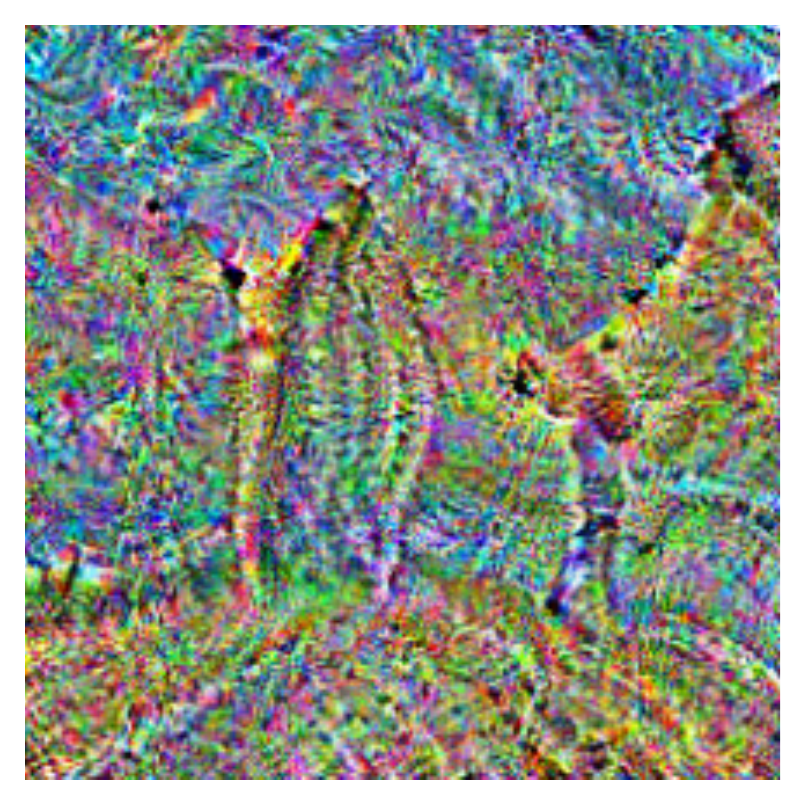}
    \end{subfigure}
    \begin{subfigure}[c]{.15\textwidth}
        \includegraphics[width=\textwidth]{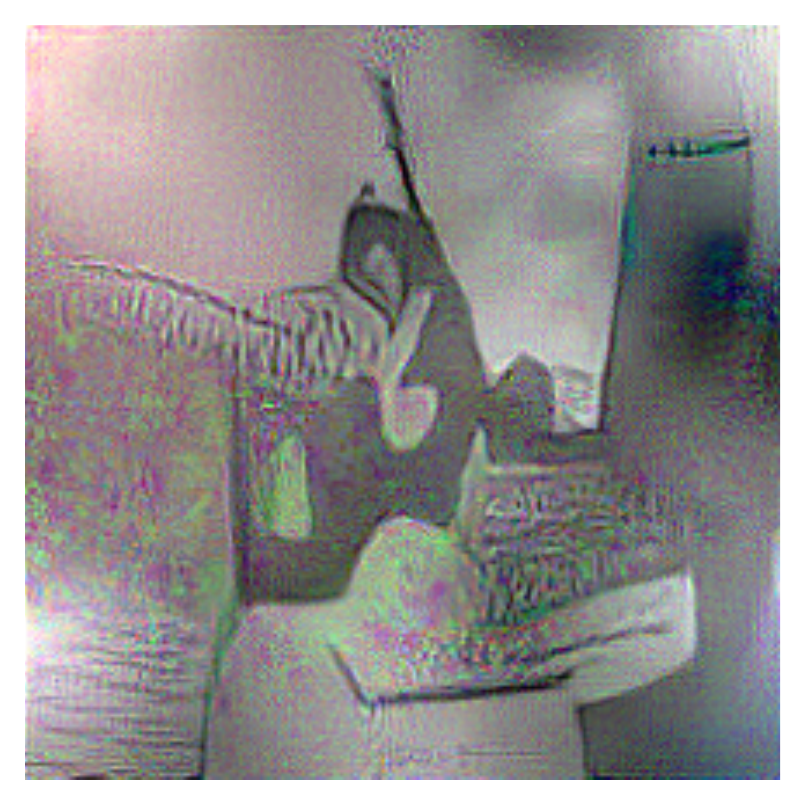}
        \includegraphics[width=\textwidth]{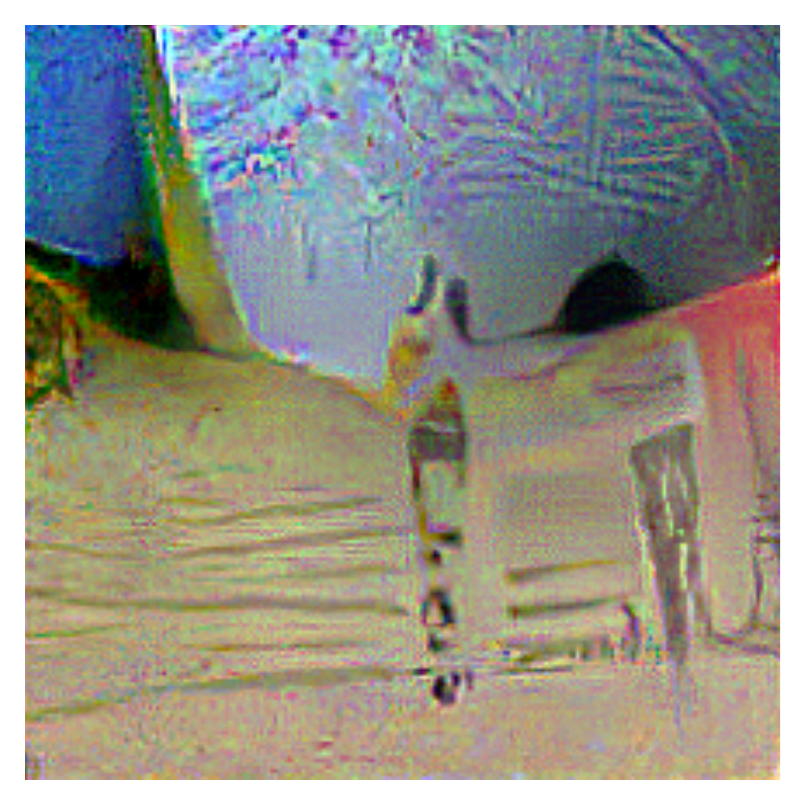}
    \end{subfigure}
    \hspace{5mm}
    \begin{subfigure}[c]{.15\textwidth}
        \includegraphics[width=\textwidth]{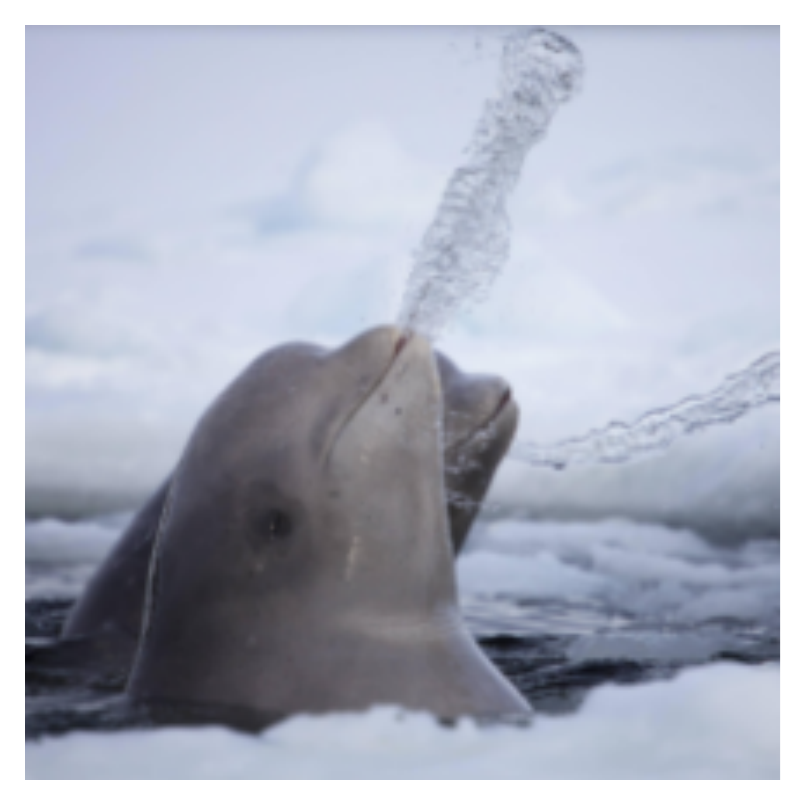}
        \includegraphics[width=\textwidth]{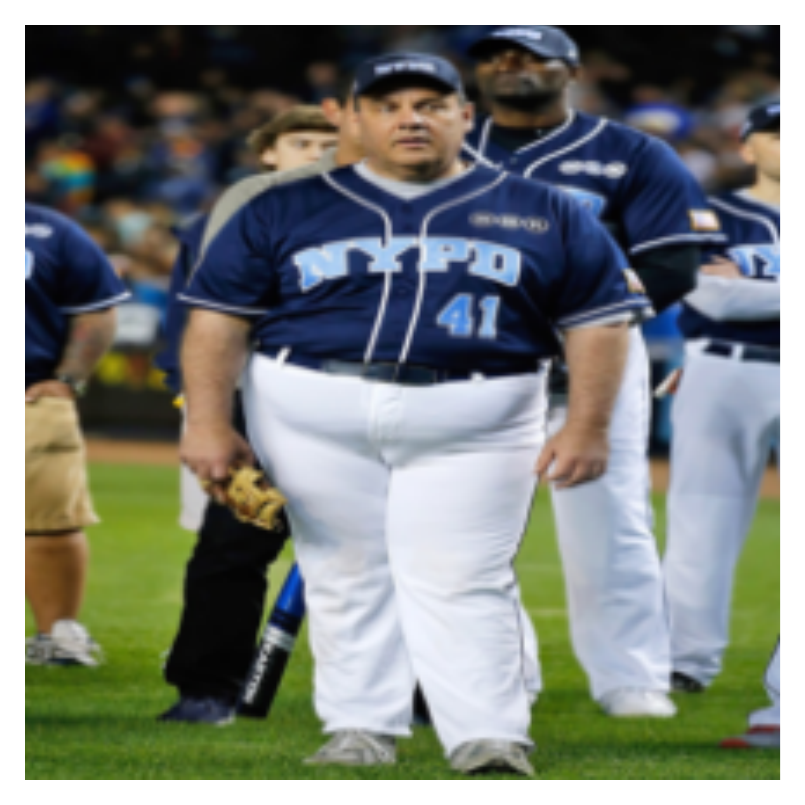}
    \end{subfigure}
    \begin{subfigure}[c]{.15\textwidth}
        \includegraphics[width=\textwidth]{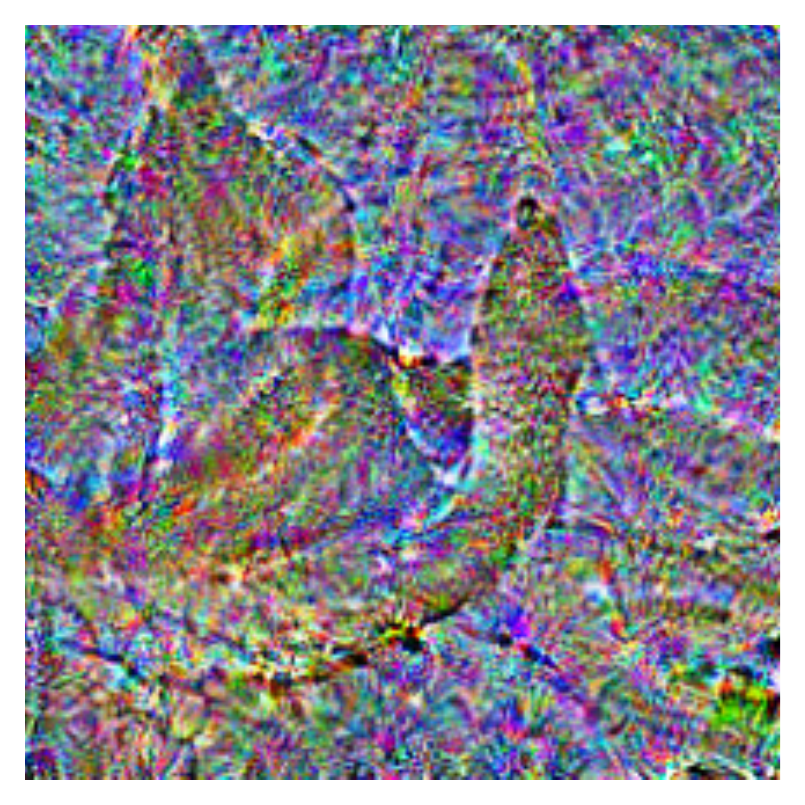}
        \includegraphics[width=\textwidth]{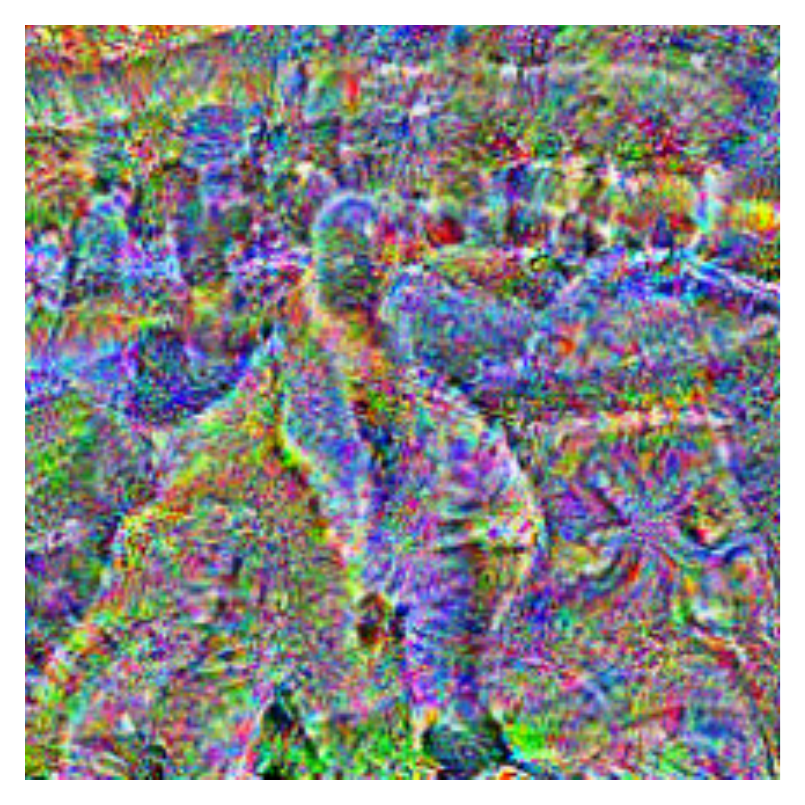}
    \end{subfigure}
    \begin{subfigure}[c]{.15\textwidth}
        \includegraphics[width=\textwidth]{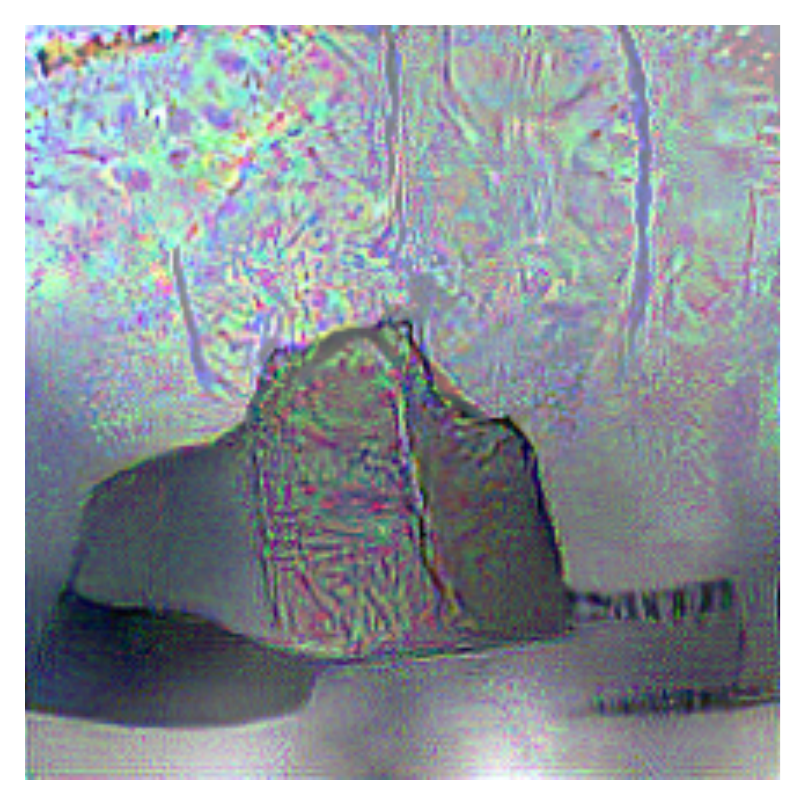}
        \includegraphics[width=\textwidth]{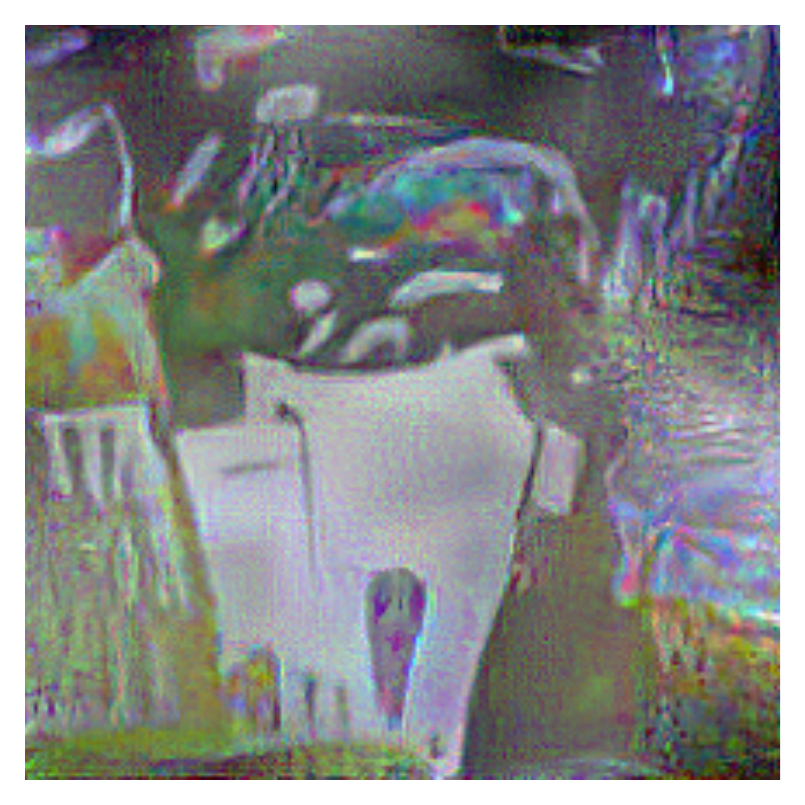}
    \end{subfigure}
    \\
    \begin{subfigure}[c]{.15\textwidth}
        \includegraphics[width=\textwidth]{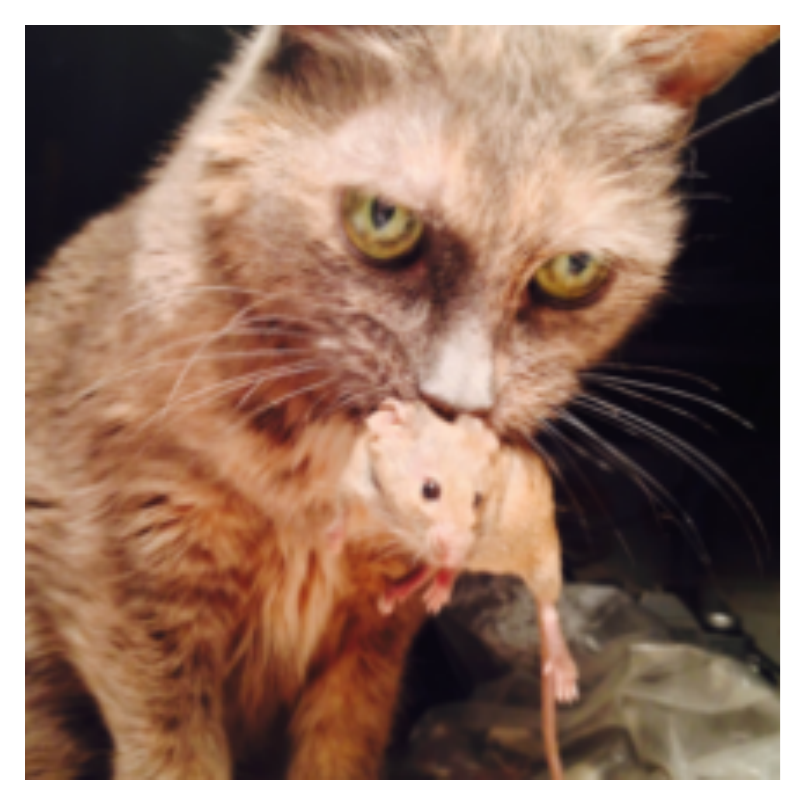}
        \includegraphics[width=\textwidth]{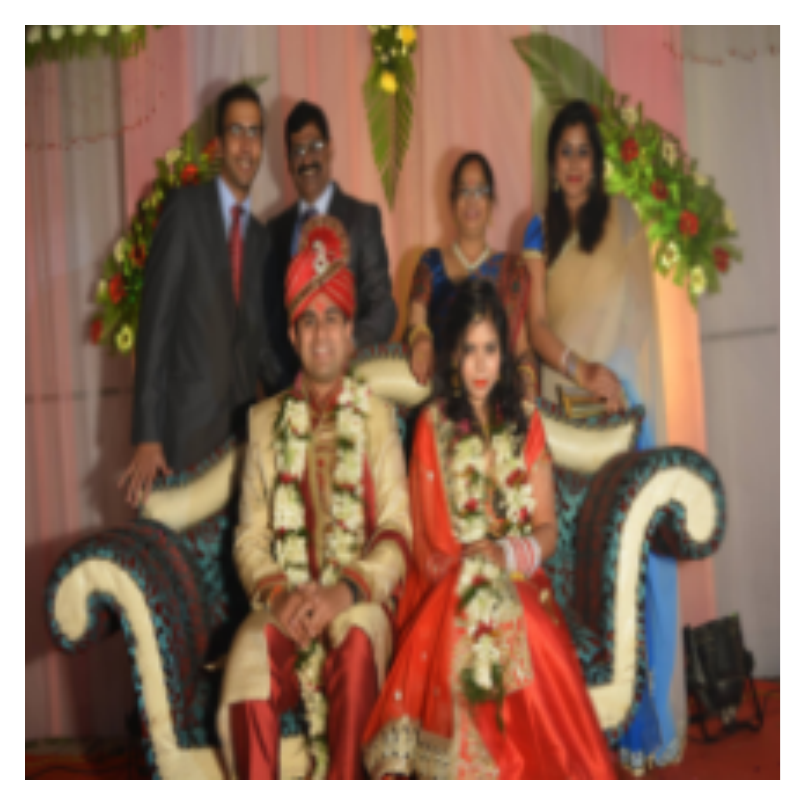}
    \end{subfigure}
    \begin{subfigure}[c]{.15\textwidth}
        \includegraphics[width=\textwidth]{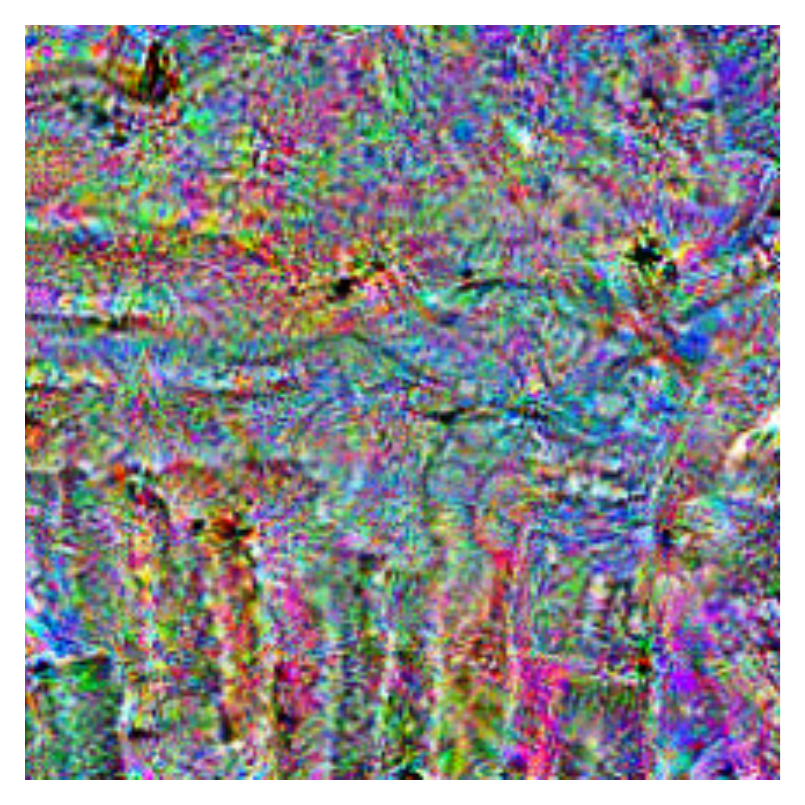}
        \includegraphics[width=\textwidth]{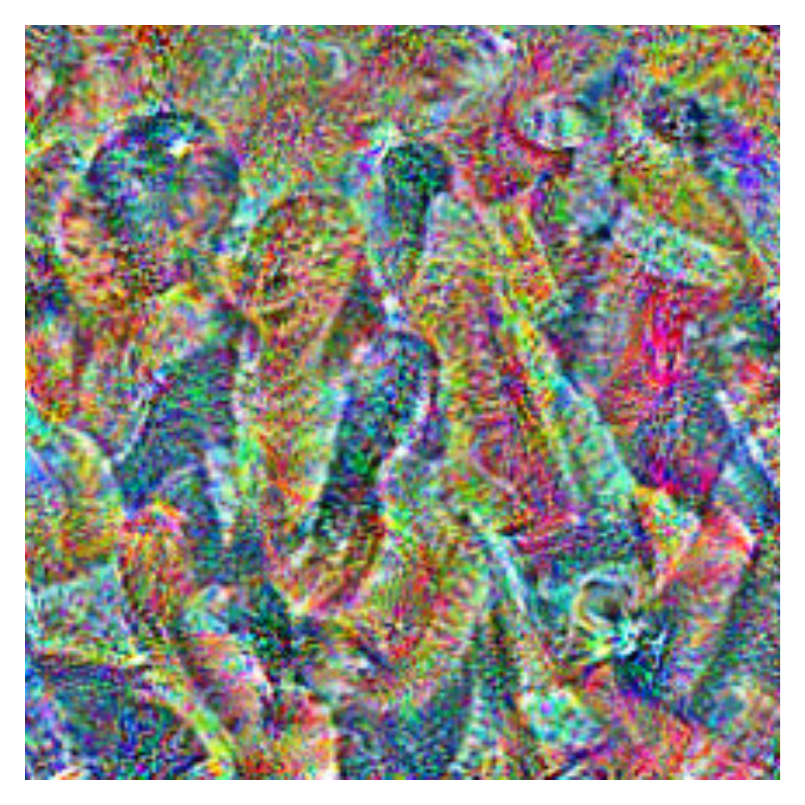}
    \end{subfigure}
    \begin{subfigure}[c]{.15\textwidth}
        \includegraphics[width=\textwidth]{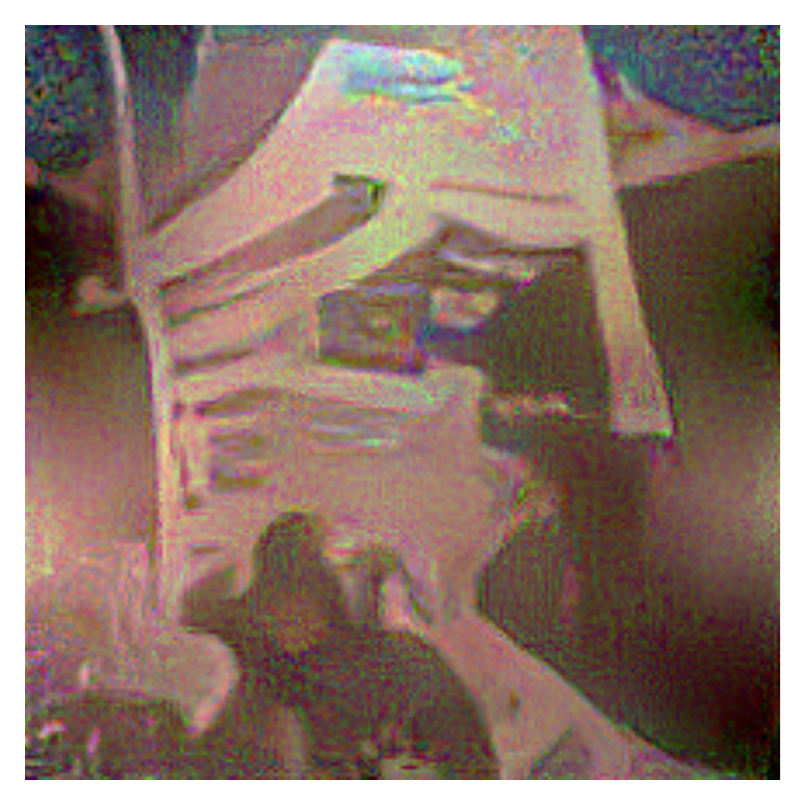}
        \includegraphics[width=\textwidth]{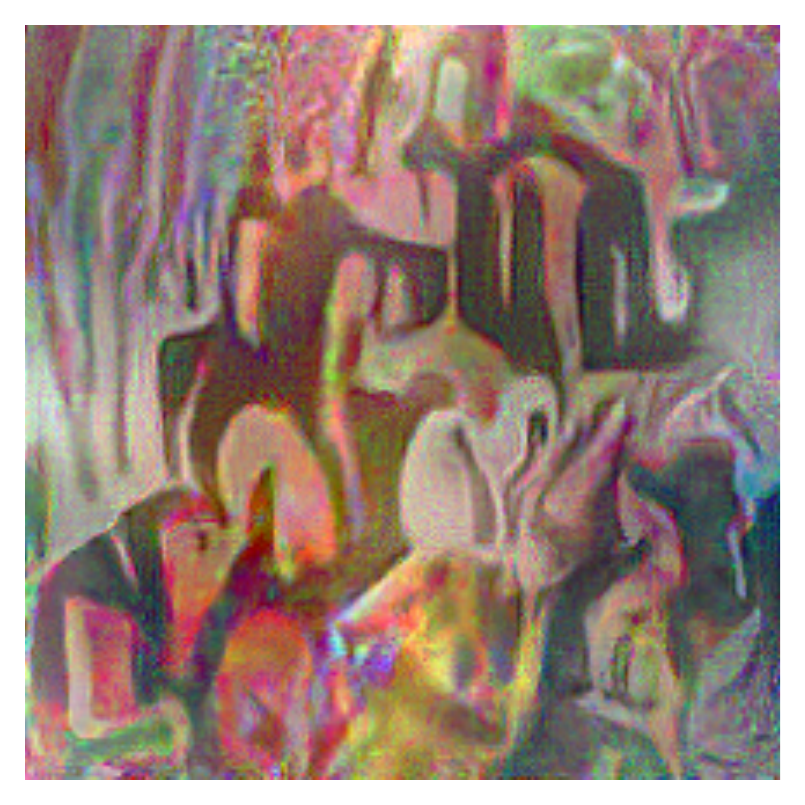}
    \end{subfigure}
    \hspace{5mm}
    \begin{subfigure}[c]{.15\textwidth}
        \includegraphics[width=\textwidth]{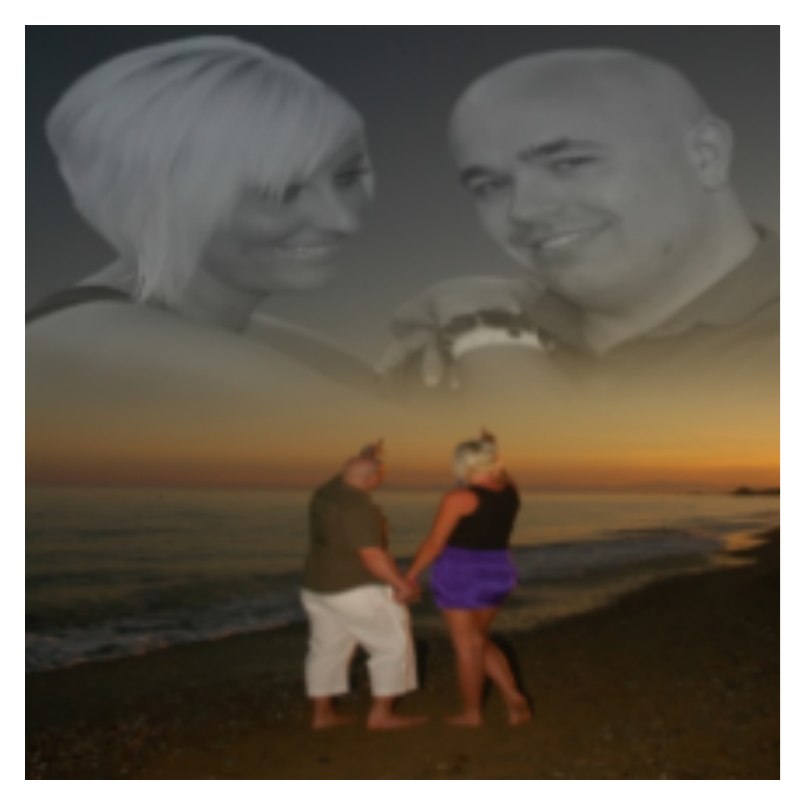}
        \includegraphics[width=\textwidth]{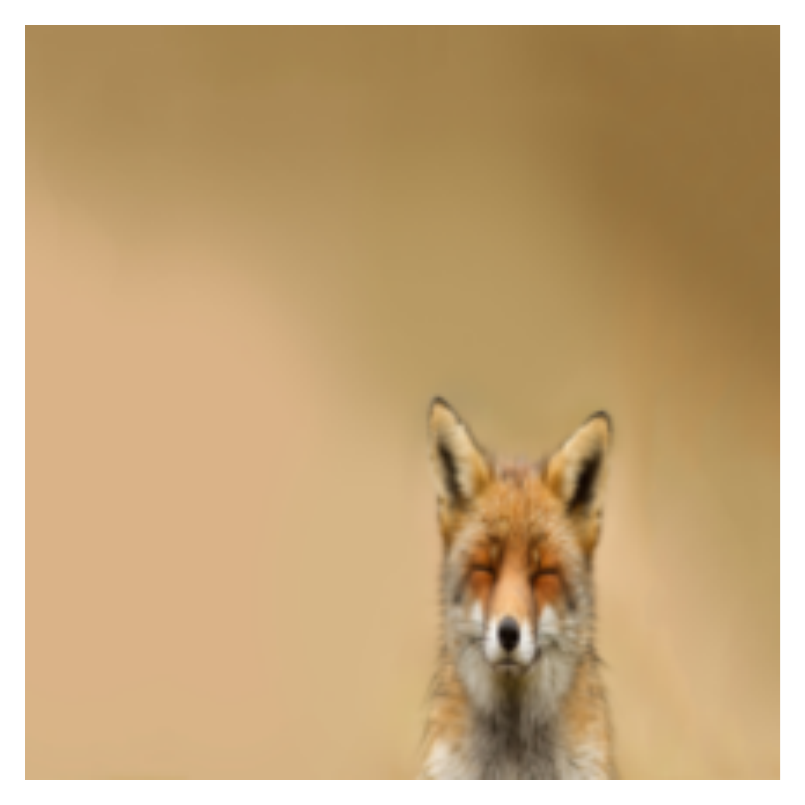}
    \end{subfigure}
    \begin{subfigure}[c]{.15\textwidth}
        \includegraphics[width=\textwidth]{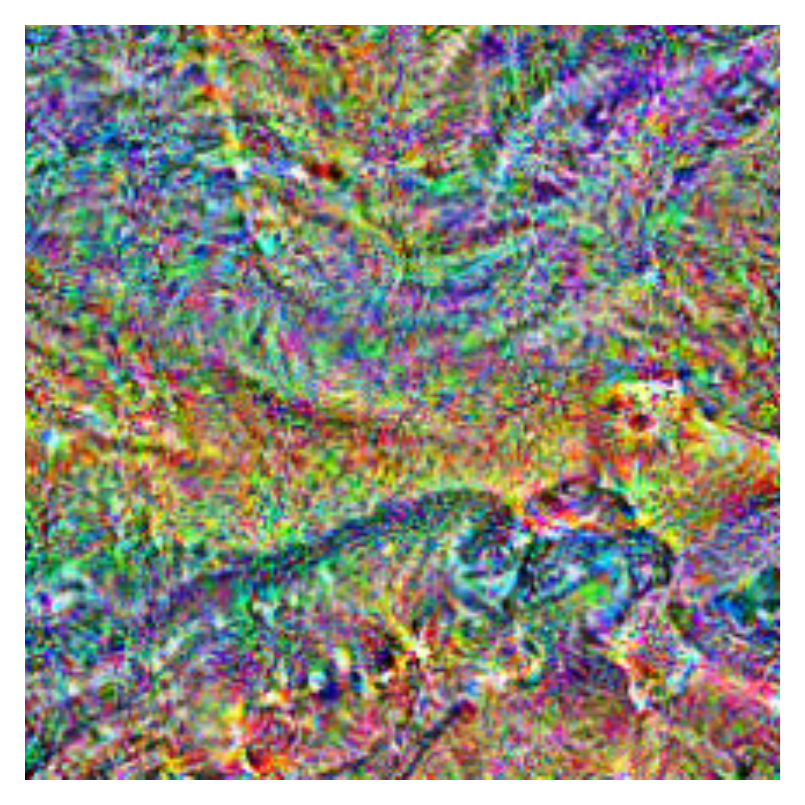}
        \includegraphics[width=\textwidth]{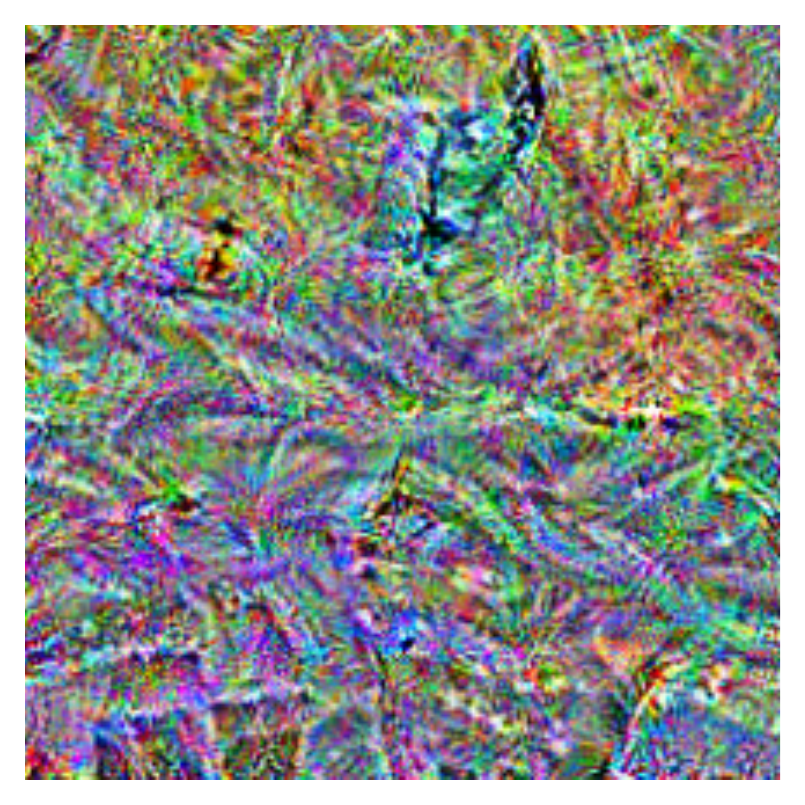}
    \end{subfigure}
    \begin{subfigure}[c]{.15\textwidth}
        \includegraphics[width=\textwidth]{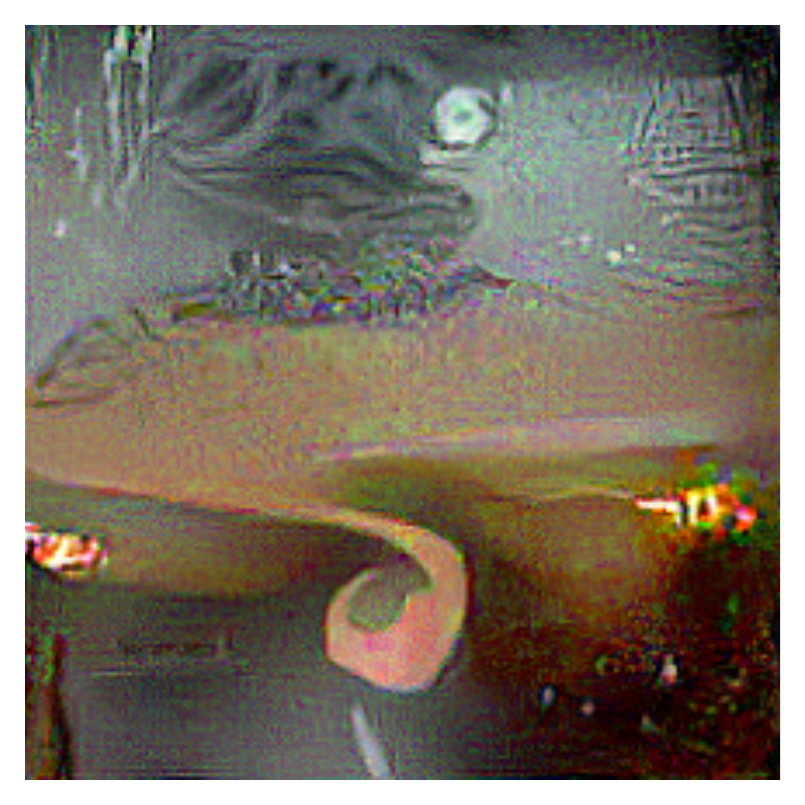}
        \includegraphics[width=\textwidth]{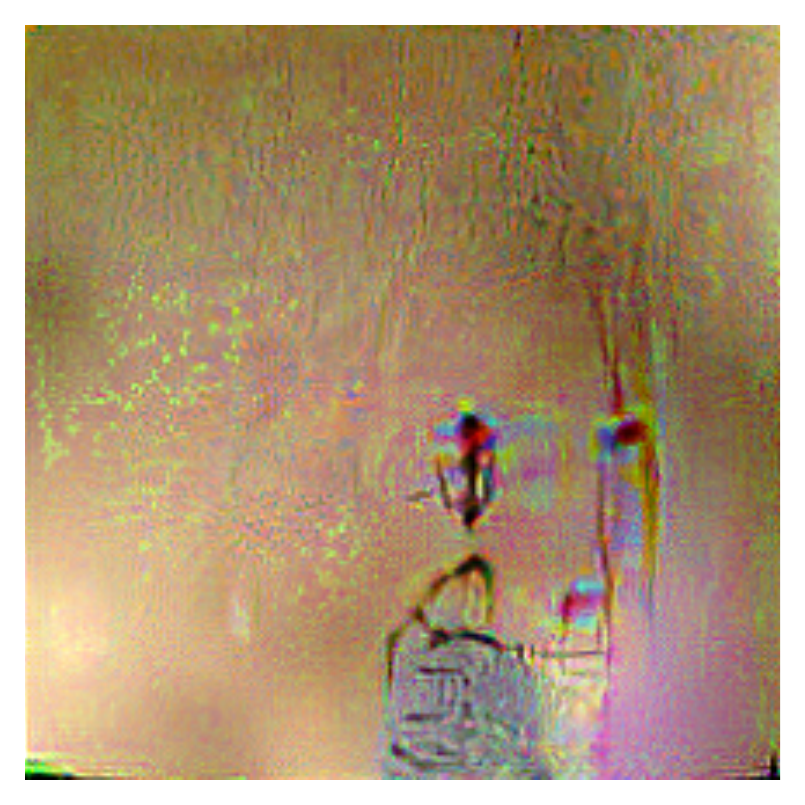}
    \end{subfigure}
    \centering
    \caption{Additional visualizations of the hash inversions $\phi^{-1}_{\approx}(x)$ for twelve original images $x$ (\textbf{left}) for a standard model (\textbf{middle}) and ARIA model with $\eps_\infty=\nicefrac{4}{255}$ (\textbf{right}), both trained on Behance1M.}
    \label{fig:hash_inversions_extra}
\end{figure*}

\paragraph{Hash inversion visualizations.}
Additional hash inversions for randomly chosen images from PSBattles can be found in Fig.~\ref{fig:hash_inversions_extra}. We can observe that in many cases hash inversions for the robust model (trained with $\eps_\infty = \nicefrac{4}{255}$) recover the shapes of original images. This is in contrast with the high-frequency noise which is observed for the standard model. 

\end{document}